\documentclass[11pt]{article}

\usepackage[final]{acl}

\usepackage{times}
\usepackage{latexsym}
\usepackage[T2A,T1]{fontenc}
\DeclareTextSymbolDefault{\cyra}{T2A}
\DeclareTextSymbolDefault{\cyre}{T2A}
\DeclareTextSymbolDefault{\cyro}{T2A}
\DeclareTextSymbolDefault{\cyrr}{T2A}
\usepackage[utf8]{inputenc}
\usepackage{microtype}
\usepackage{inconsolata}

\usepackage{amsmath}
\usepackage{amssymb}
\usepackage{bm}
\usepackage{mathtools}

\usepackage{booktabs}
\usepackage{multirow}
\usepackage{makecell}
\usepackage{tabularx}
\usepackage{array}
\usepackage[table]{xcolor}

\usepackage{graphicx}
\usepackage{subcaption}

\usepackage{algorithm}
\usepackage{algorithmicx}
\usepackage{algpseudocode}

\usepackage{xcolor}
\usepackage{xspace}

\usepackage{url}

\newcommand{\dlm}{\textsc{DLM}\xspace}
\newcommand{\ar}{\textsc{AR}\xspace}
\newcommand{\llada}{LLaDA\xspace}
\newcommand{\dream}{Dream\xspace}
\newcommand{\rdi}{\text{RDI}\xspace}
\newcommand{\ipm}{\textsc{IPM}\xspace}
\newcommand{\dgcg}{\textsc{D-GCG}\xspace}
\newcommand{\ece}{\text{ECE}\xspace}
\newcommand{\calL}{\mathcal{L}}
\newcommand{\E}{\mathbb{E}}
\newcommand{\R}{\mathbb{R}}

\newcolumntype{C}[1]{>{\centering\arraybackslash}p{#1}}

\title{Beyond the Bidirectional Promise: Re-evaluating the Robustness of Diffusion Language Models}

\author{Saurabh Yadav \\
  Microsoft \\
  \texttt{\small yadavsaurabh@microsoft.com} \\\And
  Badri Narayana Patro\\
  Microsoft \\
  \texttt{\small badripatro@microsoft.com} \\ \And
  Vijay Srinivas Agneeswaran\\
  Microsoft \\
  \texttt{\small vagneeswaran@microsoft.com} \\
  }

\begin{document}
\maketitle

\begin{abstract}
Diffusion Language Models (\dlm{}s) offer a compelling alternative to autoregressive (\ar) generation by enabling bidirectional context and iterative refinement. However, their reliability under natural input noise and adversarial attacks remains under-explored. To address this, we systematically evaluate \dlm{} robustness and calibration against \ar baselines, using two parameter-matched pairs (LLaDA-8B vs.\ LLaMA-3-8B and Dream-7B vs.\ Qwen2.5-7B) across 32 natural perturbation conditions, adversarial gradient probes, and mechanistic hidden-state analyses. This paired design effectively isolates architecture-intrinsic properties from weight-dependent behaviors. We find a nuanced robustness profile: while highly stochastic \dlm{} loss landscapes naturally resist gradient-based adversarial suffixes, they provide no guaranteed defense against natural noise, proving that everyday robustness is \emph{weight-dependent} rather than inherently architectural. Furthermore, \dlm{}s exhibit systematic overconfidence, presenting a practical deployment hazard. Most crucially, mechanistic probing reveals that all models perfectly encode input corruption ($>0.93$ linear probe accuracy), isolating behavioral fragility entirely to a \emph{decoder} routing failure. Consistent with this diagnosis, we show that surface-level prompt patching fails to improve over noisy baselines. Ultimately, \dlm{} robustness cannot be patched on; it must be fundamentally integrated into the iterative decoding loop. 
\end{abstract}
\section{Introduction}\label{sec:intro}
Diffusion Language Models (\dlm{}s) are emerging as a powerful alternative to traditional text generation. Instead of predicting tokens strictly left-to-right, models like \llada~\citep{llada2025} and \dream~\citep{dream2025} start with a fully masked sequence and iteratively reveal tokens using a discrete reverse diffusion process~\citep{d3pm2021,sedd2024,mdlm2024}. Because they process all visible tokens simultaneously through bidirectional context, they are highly effective for tasks demanding global coherence.
\begin{figure}[t]
\centering
\includegraphics[width=0.9\columnwidth]{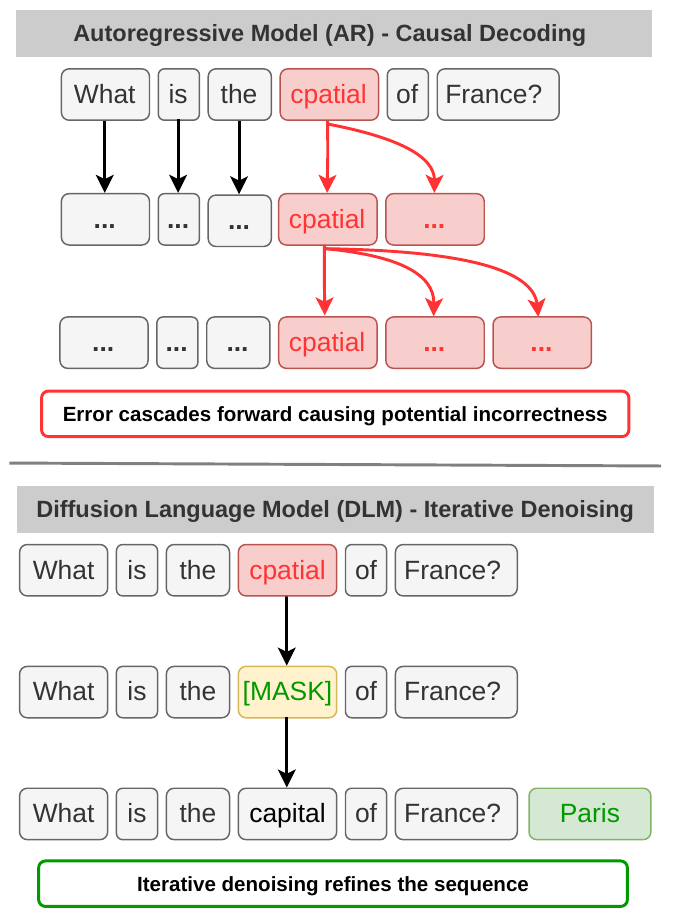}
\caption{\small Autoregressive models (top) propagate early corruptions forward via causal attention, whereas Diffusion Language Models (bottom) utilize bidirectional context to iteratively route around local noise.}
\label{fig:teaser}
\vspace{-20pt}
\end{figure}
Despite their promise, it remains unclear how well \dlm{}s handle natural input noise, such as typos or OCR mistakes. While autoregressive (\ar) models have been extensively tested under these conditions~\citep{textfooler2020,adversarial_nlp_survey,belinkov2018}, \dlm{}s remain largely unmapped territory. In theory, \dlm{}s should be more robust: \ar models rely on causal attention, where early typos can trigger cascading errors, whereas \dlm{}s process the entire input bidirectionally at every step. Yet, we do not fully understand how their iterative unmasking process interacts with noisy prompts. Furthermore, while recent studies explore \dlm{} robustness against adversarial jailbreaks~\citep{dija2026,priming2026,pad2025}, defending against benign user mistakes requires entirely different mitigation strategies.
\paragraph{Contributions:} We systematically evaluate the robustness and calibration of \dlm{}s compared to \ar models. Our core contributions are:\begin{enumerate}\item \textbf{Paired Evaluation Benchmark:} We compare two architectures (\emph{LLaDA-8B vs.\ LLaMA-3-8B} and \emph{Dream-7B vs.\ Qwen2.5-7B}) across three datasets and 32 natural noise conditions, isolating architectural traits from model-specific quirks.\item \textbf{Behavioral Robustness \& Calibration (RQ1--RQ3):} We show that robustness to natural noise is \emph{weight-dependent}, not inherently architectural (RQ1)—the diffusion objective is not a silver bullet. However, two traits replicate across both pairs: \dlm{}s are systematically overconfident compared to \ar models (RQ2), and they strongly resist gradient-based adversarial suffixes under our short-suffix budget (RQ3) (\S\ref{sec:rq1}--\S\ref{sec:rq3}).\item \textbf{Mechanistic Insights (RQ4):} Using linear probes, we find that all models (\dlm{} and \ar) detect input corruption with $>93\%$ accuracy. This indicates that \dlm{} behavioral fragility is consistent with a \emph{decoding-stage bottleneck} rather than a perceptual failure (\S\ref{sec:rq4}).\item \textbf{Falsifiable Hypothesis Testing (\ipm):} If fragility lies at decoding, patching the input prompt should not help. We test five Input Prompt Masking (\ipm) variants—none improve performance, indicating that \dlm{} robustness likely needs to be integrated directly into the generation process (\S\ref{sec:ipm}).\end{enumerate}
\section{Background}
\label{sec:background}

\subsection{Masked Diffusion Language Models}
\label{sec:background_dlm}

\dlm{}s define a forward process that corrupts clean text $\mathbf{x}_0$ by
independently masking tokens with probability $t \in [0,1]$, and learn a
reverse denoising model $p_\theta(\mathbf{x}_0 \mid \mathbf{x}_t)$ to recover
the original sequence. Following \citet{mdlm2024}, the training objective is:
\begin{equation}
  \calL_\text{mask} = \E_{t, \mathbf{m}} \Bigl[
    -\sum_{i:\, m_i=1} \log p_\theta(x_i \mid \mathbf{x}_{\backslash m},\, t)
  \Bigr]
  \label{eq:loss}
\end{equation}
where $\mathbf{m} \sim \text{Bernoulli}(t)^L$ is the binary mask. The key
architectural choice is a \emph{bidirectional} transformer with no causal mask,
enabling each position to attend to all others. During inference, generation
proceeds by initialising the response as $[\textsc{mask}, \ldots, \textsc{mask}]$
and iteratively unmasking the most confident tokens over $T$ steps.

\paragraph{Critical architecture detail.}
In the inference loop, the \emph{prompt tokens are fixed} and never masked.
The reverse diffusion process acts only on the response tokens. This means
the model does \emph{not} automatically correct noisy input tokens --- it
conditions on them unchanged. This is the central motivation for \ipm{}.

\subsection{Related Work}
\label{sec:related}

\paragraph{DLM adversarial safety.}
Since early 2025, a body of work has studied intentional adversarial attacks
on \dlm{}s: DIJA~\citep{dija2026} achieves near-100\% jailbreak success via
interleaved mask-text injections; DiffuGuard~\citep{diffuguard2026} proposes
training-free defences; \citet{priming2026} exploit affirmative token injection;
and TrajHijack~\citep{trajhijack2026} manipulates the denoising trajectory.
\citet{gcgdlm2025} adapted the GCG attack~\citep{zou2023universal} to \dlm{}s with
prefix and random suffixes. Our work is \textbf{complementary}: we study
benign natural noise, not intentional attacks, and propose \ipm{} as a
constructive remedy rather than a vulnerability analysis.

\paragraph{Robustness of AR LLMs to noise.}
Natural robustness of \ar models has been studied in terms of typos
\citep{belinkov2018}, and
adversarial word substitutions~\citep{textfooler2020,adversarial_nlp_survey}.
No analogous study exists for \dlm{}s.

\paragraph{Masked LM as noise corrector.}
\citet{masked_lm_generation} showed that masked language models can fill in
corrupted positions at inference time (Mask-Predict). We extend this insight
to modern \dlm{}s operating as generative models, proposing \ipm{} as a
\emph{pre-generation} denoising stage.

\section{Problem Formulation}
\label{sec:problem}

Let $\mathcal{M}$ denote a \dlm or \ar language model. For a clean prompt $x$, let $\tilde{x} = \phi(x, r)$ be a perturbed prompt generated by noise function $\phi$ at corruption rate $r \in [0,1]$. We structure our evaluation around four distinct research questions to systematically unpack both model behavior and internal mechanics:
\begin{itemize}
  \item \textbf{RQ1} (\S\ref{sec:rq1}): How does task accuracy degrade as natural noise $r$ increases, and do \dlm{}s and \ar models differ systematically?
  \item \textbf{RQ2} (\S\ref{sec:rq2}): How does input noise impact confidence calibration, and are \dlm{}s more susceptible to overconfidence?
  \item \textbf{RQ3} (\S\ref{sec:rq3}): How do the architectures compare in their resistance to adversarial suffixes, and what do gradient probes reveal about their underlying loss landscapes?
  \item \textbf{RQ4} (\S\ref{sec:rq4}): What mechanistic factors within the internal hidden states explain the observed behavioral differences?
\end{itemize}

\subsection{Models and Datasets}
\label{subsec:models_benchmarks}

We evaluate two parameter-matched \dlm--\ar pairs: \textbf{LLaDA-8B-Instruct}~\citep{llada2025} vs.\ \textbf{LLaMA-3-8B-Instruct}~\citep{llama3_2024}, and \textbf{Dream-7B-Instruct}~\citep{dream2025} vs.\ \textbf{Qwen2.5-7B-Instruct}~\citep{qwen25_2024}. We test three core cognitive capabilities: factual recall (TriviaQA~\citep{triviaqa2017}), mathematical reasoning (GSM8K~\citep{gsm8k2021}), and science QA (ARC-Challenge~\citep{arc2018}). 

For \dlm inference, we use the original authors' recommended generation recipes: semi-autoregressive block decoding for LLaDA (block length 32, low-confidence remasking, $T{=}128$) and entropy-ranked confidence remasking for Dream ($T{=}512$, $\tau{=}0.2$, top-$p{=}0.95$). Our proposed \ipm pre-processing uses $T_d{=}64$ denoising steps.

\subsection{Perturbation Taxonomy}
\label{subsec:perturbations}

We simulate natural input noise using nine deterministic perturbations across two levels, providing a comprehensive stress test. \emph{Character-level (5):} transpositions (e.g., \textit{teh} $\to$ \textit{the}), random deletions, random insertions, keyboard adjacencies (QWERTY substitutions), and homoglyphs (Unicode lookalikes, e.g., Cyrillic `а'). Evaluated at $r \in \{0.05, 0.10, 0.20, 0.30\}$. \emph{Word-level (4):} random content word deletion, local shuffling (window size 3), synonym substitution, and word repetition. Evaluated at $r \in \{0.10, 0.20, 0.30\}$.

This taxonomy yields $5{\times}4 + 4{\times}3 = 32$ distinct noise conditions per benchmark.

\subsection{Evaluation Metrics}
\label{subsec:metrics}

To quantify performance drops, we define the \emph{Robustness Degradation Index} (\rdi) at noise rate $r$:
\begin{equation}
  \rdi(r) = \frac{A_\text{clean} - A_r}{A_\text{clean}}
  \label{eq:rdi}
\end{equation}
where $A_\text{clean}$ is baseline clean accuracy and $A_r$ is accuracy at rate $r$. $\rdi \in [0,1]$, where $0$ indicates no degradation. For cross-model comparisons, we compute the Area Under the RDI curve (AURDI):
\begin{equation}
  \text{AURDI} \approx \sum_r \rdi(r) \cdot \Delta r
  \label{eq:aurdi}
\end{equation}

For calibration, we calculate Expected Calibration Error (\ece) using 15 equal-width bins~\citep{calibration_guo2017}. \dlm confidence is the geometric mean of the per-token \emph{maximum} softmax probabilities from a single bidirectional pass over the final committed sequence $\hat{\mathbf{y}}$ (not the chosen-token probability at commitment time):
\begin{equation}
  c_\text{DLM}(\hat{\mathbf{y}}) = \exp\!\Bigl(
    \tfrac{1}{L}\sum_{i=1}^L \log p_\theta(\hat{y}_i \mid \hat{\mathbf{y}}_{\backslash i})
  \Bigr)
  \label{eq:dlm_conf}
\end{equation}
The \ar score uses the geometric mean of chosen-token probabilities under causal masking: the aggregation is identical across architectures, while the per-token basis differs. We ablate this definition (chosen/commit-time basis; mean/min aggregation) in Appendix~\ref{app:calib_ablation}.
\section{D-GCG Robustness Probe}
\label{sec:dgcg}

While natural noise evaluates everyday robustness, adversarial attacks reveal a model's true worst-case vulnerabilities. However, standard gradient-based attacks like the Greedy Coordinate Gradient (GCG)~\citep{zou2023universal} are built for \ar models, relying on deterministic, left-to-right gradient signals. \dlm{}s, which generate text iteratively over a learned reverse diffusion process, are structurally incompatible with this approach. 

To bridge this gap and probe the \dlm gradient landscape, we introduce \textbf{Diffusion-GCG} (\dgcg{}).

\subsection{Formulation}
\label{subsec:dgcg_form}

Given a prompt $x$ and a target response $y$, we aim to optimize an adversarial suffix $\mathbf{s} \in \mathcal{V}^K$ that maximizes the model's difficulty in predicting the target:
\begin{multline}
  \mathbf{s}^* = \arg\max_{\mathbf{s}} \; \E_{t \sim \mathcal{U}(0,1),\; \mathbf{m} \sim \text{Bern}(t)} \\
  \Bigl[ -\!\sum_{i:\, m_i=1} \log p_\theta(y_i \mid \mathbf{y}_{\backslash m},\, [x;\mathbf{s}],\, t) \Bigr]
  \label{eq:dgcg}
\end{multline}
Backpropagating through all $T$ denoising steps would be computationally prohibitive and easily exceed standard VRAM limits. Instead, we estimate gradients using a single sampled timestep $t$ and mask $\mathbf{m}$:
\begin{equation}
\nabla_{\mathbf{s}} \approx
\nabla_{\mathbf{s}} \Bigl[
-\!\sum_{i:\, m_i=1} \log p_\theta(y_i \mid \mathbf{y}_{\backslash m},\, [x;\mathbf{s}],\, t)
\Bigr]
\label{eq:dgcg_grad}
\end{equation}

This single-step unbiased estimator mirrors standard diffusion training and keeps memory usage feasible ($\approx 1.5\times$ standard inference). Conceptually, this acts as \emph{gradient trajectory sampling}. Because a \dlm{}'s loss surface evolves continuously across $T$ steps, sampling at a single timestep maps the local landscape at one specific point in the trajectory. Aggregating these gradients across optimization steps allows us to measure the overall coherence or incoherence of the loss landscape. We utilize \dgcg{} strictly as an analytical probe to uncover these hidden dynamics.

\section{Mechanistic Probing Methodology}
\label{sec:mech_methodology}

When a model fails on a noisy prompt, behavioral metrics (like accuracy degradation) cannot tell us \emph{why}. The failure could stem from the encoder (the model fails to represent the corrupted token properly) or the decoder (the model recognizes the noise but fails to suppress it during generation). To isolate this bottleneck and investigate the mechanisms driving behavioral robustness (RQ4), we employ three targeted probing techniques:

\paragraph{Token-Level Linear Probing.}
Our primary diagnostic tool tests whether the encoder explicitly perceives input corruption. We extract hidden-state vectors from 17 evenly spaced layers (layers $\{0,2,4,\ldots,30,31\}$ of $32$) for perturbed inputs. We then train an $\ell_2$-regularized logistic regression (\texttt{C}=1, \texttt{max\_iter}=500) on a balanced 80/20 train/test split to classify tokens as either corrupted ($1$) or clean ($0$). 

Our crucial contribution here is the \emph{linearity} of the classifier. Because a complex, non-linear transformer block could theoretically extract any feature its activations support, achieving high accuracy with a simple linear probe indicates that the feature is already cleanly separable in the model's embedding space. If the linear probe succeeds, the encoder is not blind to the noise, so any downstream failure is consistent with a decoding-stage bottleneck rather than an encoding failure. We report the maximum test accuracy and AUROC across the sampled layers and note the layer index at which the maximum occurs.

\paragraph{Denoising Trajectory Analysis (\dlm{}s only).}
Because \dlm{}s refine text iteratively, we can trace exactly when and how corruption derails the generation pathway. We compare token-prediction trajectories for clean versus perturbed inputs across eight logarithmically spaced snapshot steps (from $t{=}T$ down to $t{\approx}0.01T$). We measure three key metrics: (i) \emph{ID divergence} (the exact fraction of steps where noisy and clean predictions split), (ii) \emph{confidence drop} (mean reduction in prediction confidence), and (iii) \emph{stability drop} (mean reduction in trajectory consistency). This maps the lifecycle of a corruption as it propagates through the diffusion process.

\paragraph{Attention Pattern Analysis.}
To understand how \ar models dynamically react to noise, we use forward hooks to compute the \emph{corruption attention ratio}---the fraction of total attention mass allocated specifically to corrupted-token positions, averaged across heads and output positions. A ratio $> 1.0$ indicates the model is actively concentrating attention on the noise (e.g., compensatory redistribution), while a ratio $< 1.0$ indicates active suppression. Contrasting these \ar routing strategies against \dlm trajectory behaviors allows us to pinpoint architectural differences in how noise is managed.

\section{Experiments}
\label{sec:experiments}

\subsection{Experimental Setup}
\label{sec:exp_setup}

We conducted all experiments on two NVIDIA RTX 4090 GPUs, which provided sufficient memory to evaluate models in native precision without quantization. We used PyTorch 2.11, CUDA 13.0, and Transformers 4.45.2. To account for perturbation variance, we report the mean performance across three random seeds. Unless otherwise noted, each perturbation condition is evaluated on $n{=}200$ samples; mechanistic probing uses $n{=}100$ sequences and the D-GCG probe uses $n{=}50$. Character- and word-level perturbations are applied to the \emph{entire formatted prompt} (instructions, question, and any answer choices), identically for all four models. Exact prompt templates and hyperparameters are detailed in Appendices~\ref{app:prompts} and~\ref{app:hyperparams}.

\subsection{RQ1: Natural Robustness Degradation}
\label{sec:rq1}

\begin{figure}[t]
\centering
\includegraphics[width=\columnwidth]{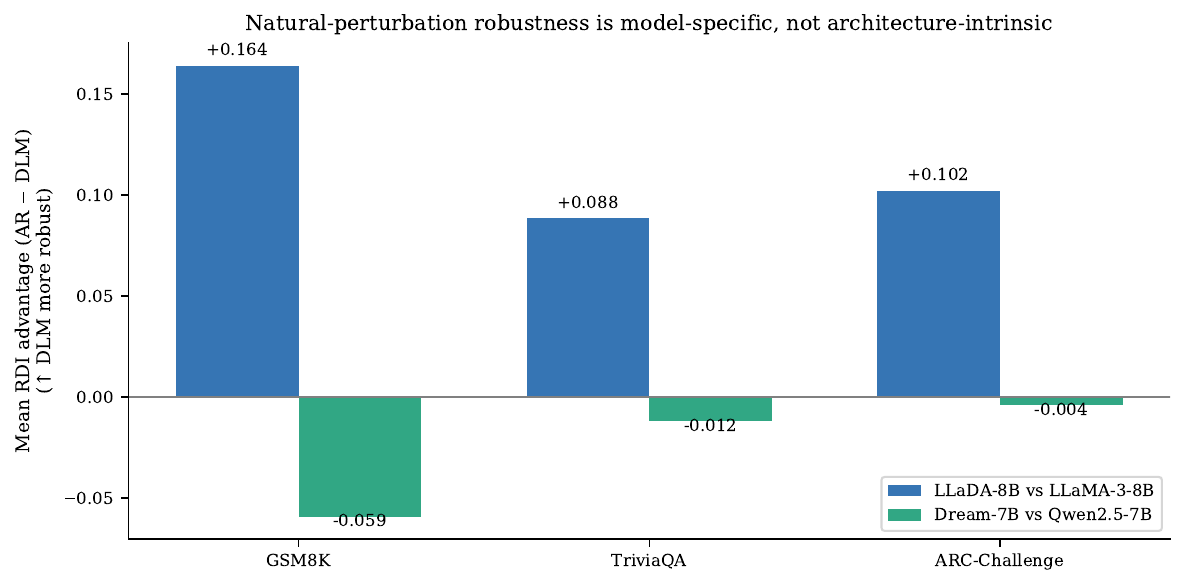}
\caption{\small Mean RDI advantage across 32 noise conditions. LLaDA consistently outperforms LLaMA-3, while Dream's advantage over Qwen2.5 varies by task, demonstrating that natural robustness is weight-dependent, not inherently architectural.}
\label{fig:rdi_crosspair}
\vspace{-10pt}
\end{figure}

\begin{figure*}[t]
\centering
\includegraphics[width=\textwidth]{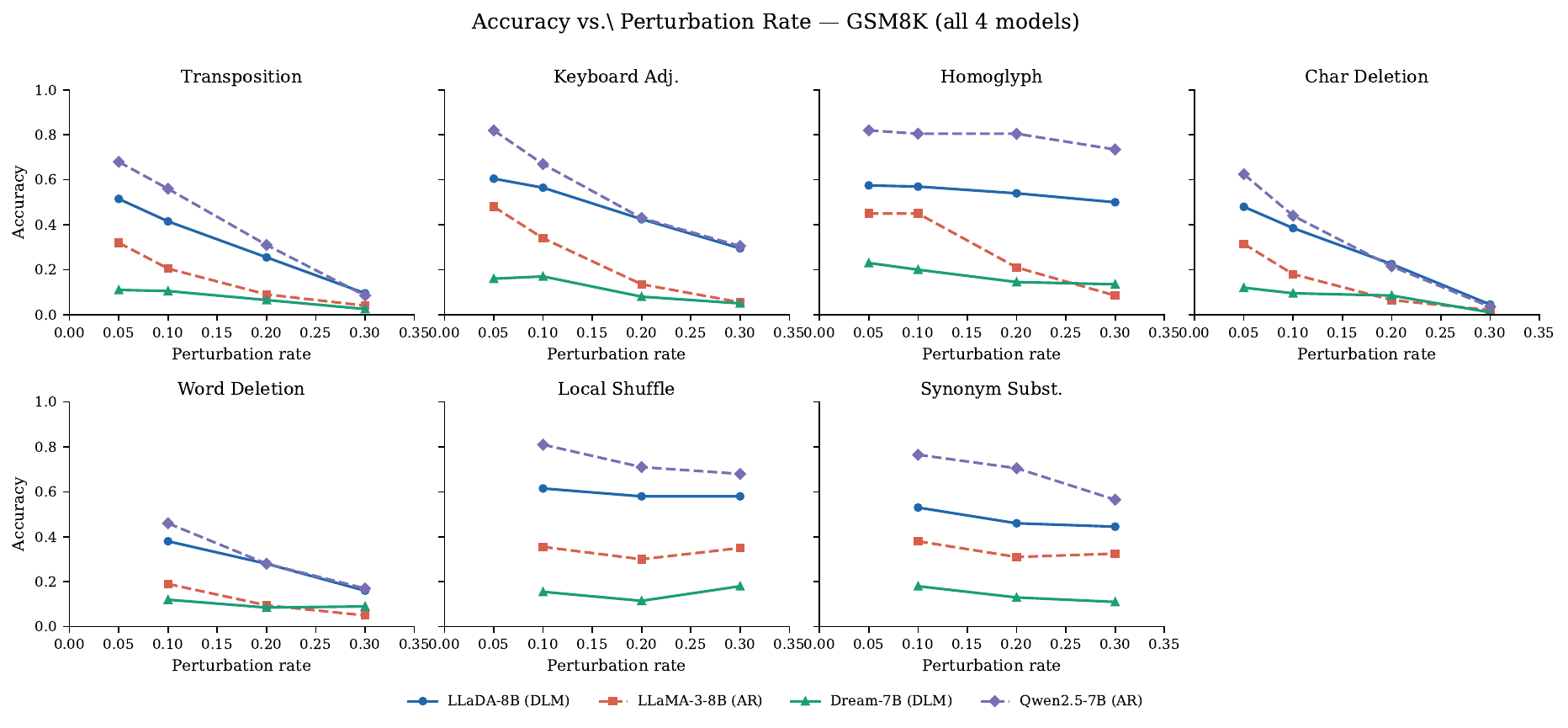}
\caption{\small Accuracy vs. perturbation rate on GSM8K. Both DLMs (solid lines) maintain higher relative accuracy under noise compared to their AR counterparts (dashed lines)}
\label{fig:sweep_gsm8k}
\end{figure*}

Table~\ref{tab:rq1_main} reports clean baseline accuracy, mean \rdi, and the \dlm advantage rate across 32 conditions. Figure~\ref{fig:rdi_crosspair} summarizes cross-pair divergence, while Figure~\ref{fig:sweep_gsm8k} details GSM8K degradation curves (full heatmap in Appendix~\ref{app:full_results}).

\begin{table}[t]
\centering
\small
\caption{\small Clean accuracy and natural-perturbation robustness (mean over 32 conditions, $n{=}200$ each). Bold marks the within-pair winner. Acc$_\text{clean}$: clean accuracy. Acc$_r$: mean accuracy under noise. $\overline{\text{RDI}}$: mean relative degradation (lower is better). Adv.\%: conditions where the \dlm beats its \ar counterpart. RDI is \emph{relative} and should be read jointly with Acc$_r$ (e.g., LLaDA's low TriviaQA RDI coexists with low absolute accuracy). Paired significance in App.~\ref{app:significance}.}
\label{tab:rq1_main}
\setlength{\tabcolsep}{3pt}
\resizebox{\columnwidth}{!}{%
\begin{tabular}{l l C{1.1cm} C{1.1cm} C{1.1cm} C{1.0cm}}
\toprule
\textbf{Task} & \textbf{Model} & \textbf{Acc$_\text{clean}$} & \textbf{Acc$_r$} & $\overline{\text{RDI}}$ & \textbf{Adv.\%} \\
\midrule
\multirow{4}{*}{TriviaQA}
  & LLaDA-8B    & 0.185 & 0.128 & \cellcolor{blue!15}\textbf{0.319} & \multirow{2}{*}{\textbf{88\%}} \\
  & LLaMA-3-8B  & \cellcolor{blue!15}0.545 & 0.323 & 0.408 & \\
  \cmidrule(lr){2-6}
  & Dream-7B    & 0.220 & 0.144 & 0.345 & \multirow{2}{*}{41\%} \\
  & Qwen2.5-7B  & \cellcolor{blue!5}0.400 & 0.267 & \cellcolor{blue!5}\textbf{0.333} & \\
\midrule
\multirow{4}{*}{GSM8K}
  & LLaDA-8B    & \cellcolor{blue!15}0.625 & 0.434 & \cellcolor{blue!15}\textbf{0.306} & \multirow{2}{*}{\textbf{91\%}} \\
  & LLaMA-3-8B  & 0.460 & 0.245 & 0.470 & \\
  \cmidrule(lr){2-6}
  & Dream-7B    & 0.190 & 0.121 & 0.374 & \multirow{2}{*}{34\%} \\
  & Qwen2.5-7B  & \cellcolor{blue!5}0.825 & 0.568 & \cellcolor{blue!5}\textbf{0.315} & \\
\midrule
\multirow{4}{*}{ARC-Chall.}
  & LLaDA-8B    & \cellcolor{blue!15}0.860 & 0.643 & \cellcolor{blue!15}\textbf{0.252} & \multirow{2}{*}{\textbf{88\%}} \\
  & LLaMA-3-8B  & 0.840 & 0.542 & 0.354 & \\
  \cmidrule(lr){2-6}
  & Dream-7B    & 0.815 & 0.649 & 0.204 & \multirow{2}{*}{47\%} \\
  & Qwen2.5-7B  & \cellcolor{blue!5}0.900 & 0.720 & \cellcolor{blue!5}\textbf{0.200} & \\
\bottomrule
\end{tabular}%
}
\end{table}

\paragraph{Weight-Dependent, Not Architecture-Intrinsic:}
Robustness to natural noise is not a guaranteed feature of the diffusion objective. LLaDA-8B is more robust than LLaMA-3-8B across 88--91\% of conditions, and this advantage is statistically significant on all three tasks (paired bootstrap 95\% CIs exclude zero; Wilcoxon $p<0.001$; App.~\ref{app:significance}). The Dream-7B versus Qwen2.5-7B comparison does \emph{not} replicate this pattern: Dream surpasses Qwen2.5 on none of the three tasks---its RDI advantage is statistically indistinguishable from zero on TriviaQA ($p{=}0.49$) and ARC ($p{=}0.66$) and significantly \emph{negative} on GSM8K ($p{=}0.013$). This indicates robustness is tied to specific pretraining data and weights, so \dlm robustness claims should be validated on a per-model basis.

\paragraph{Baseline Disparities Are Knowledge, Not Decoding:}
Dream's lower clean accuracy on TriviaQA (0.220 vs.\ 0.400 for Qwen) occurs because it generates fluent but factually incorrect answers (e.g., confidently stating a celebrity is a Virgo instead of a Scorpio). This points to weaker factual coverage in its pretraining data rather than a decoding failure. Because our \rdi metric normalizes by clean accuracy, this baseline gap does not skew the robustness comparison.

\paragraph{Consistent Trends Across Architectures:}
Three patterns hold across all models: (i) character-level noise degrades performance more heavily than word-level noise; (ii) \ar models handle homoglyph substitutions best, as modern tokenizers automatically map these look-alikes to their Latin equivalents; and (iii) ARC-Challenge shows the least cross-model variance, likely due to its constrained multiple-choice format.

\subsection{RQ2: Calibration Under Noise}
\label{sec:rq2}

\begin{figure}[t]
\centering
\includegraphics[width=\columnwidth]{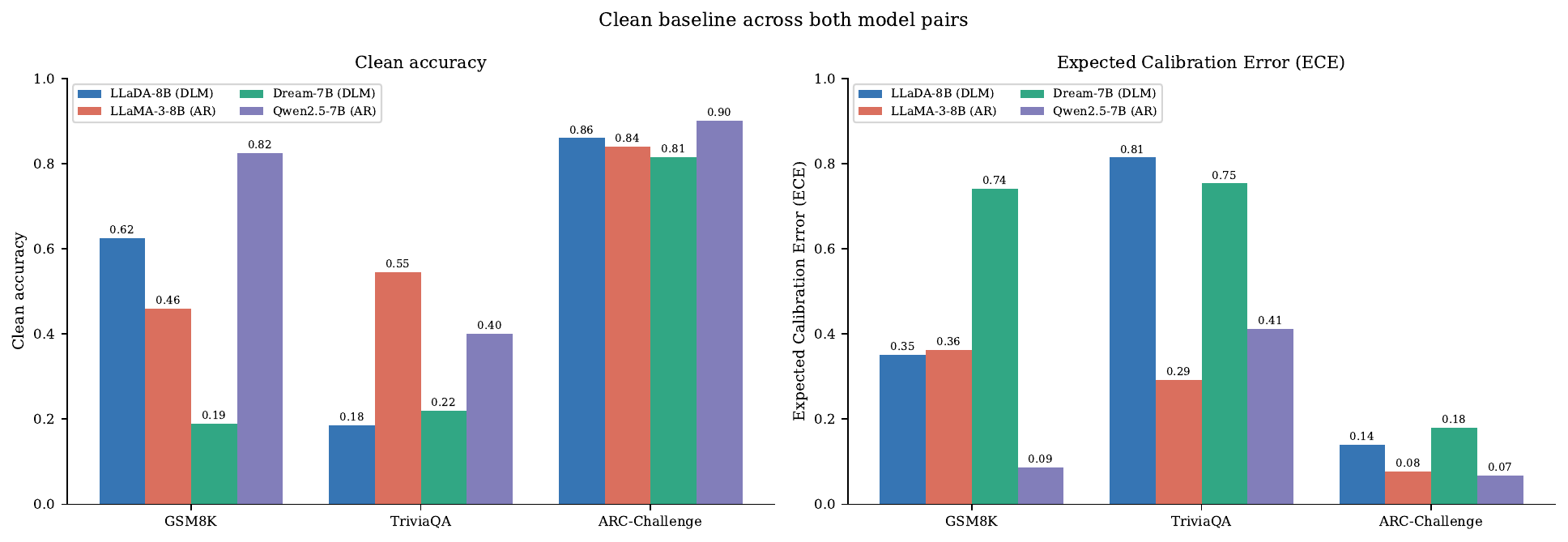}
\caption{\small Clean baseline accuracy (left) and Expected Calibration Error (ECE) (right). DLMs exhibit systematically higher ECE, indicating persistent overconfidence compared to AR models.}
\label{fig:baseline_crosspair}
\vspace{-10pt}
\end{figure}

\begin{figure*}[t]
\centering
\includegraphics[width=0.98\textwidth]{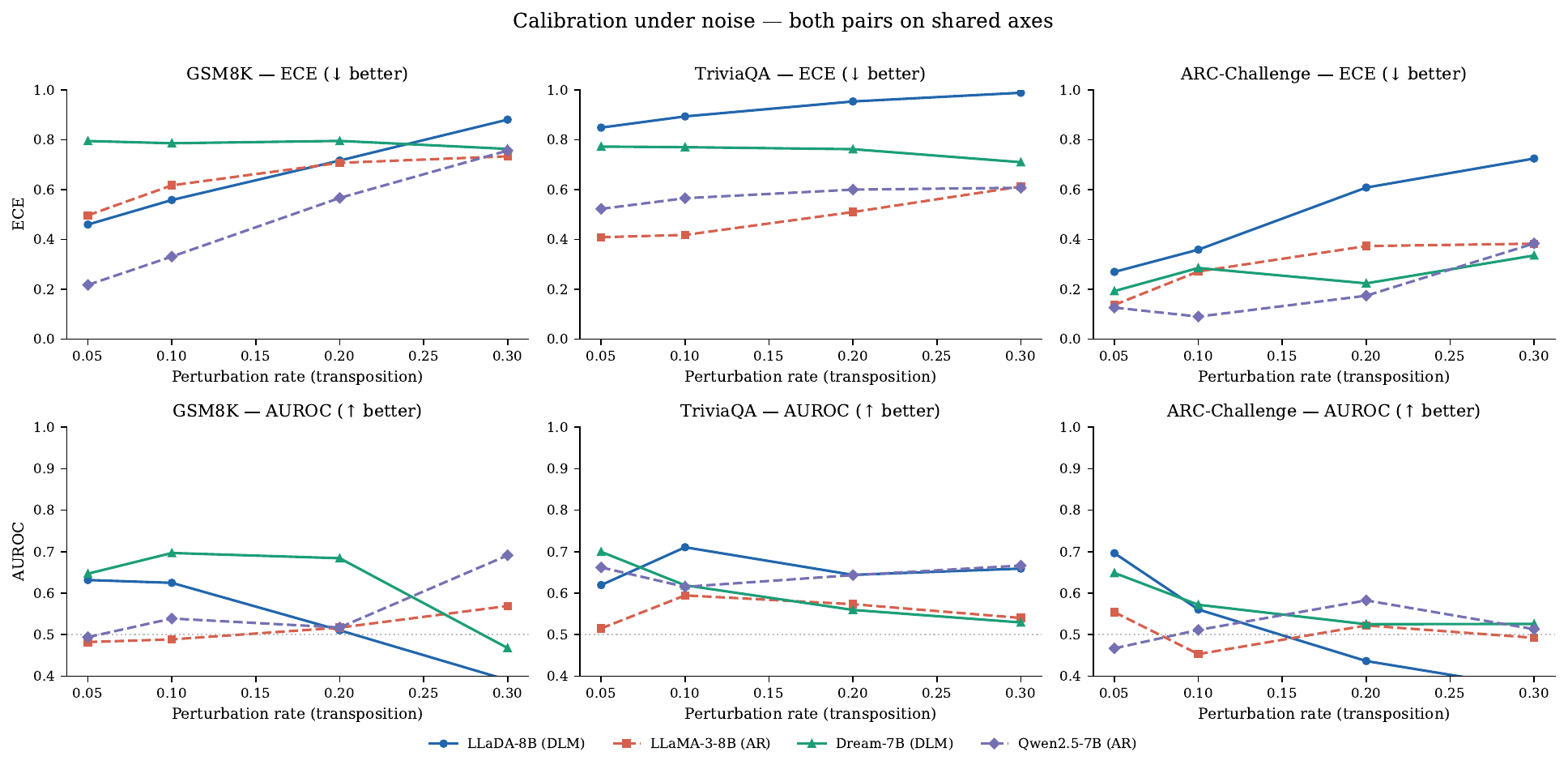}
\caption{\small ECE and AUROC vs. transposition rate for the LLaDA pair and Dream pair. DLMs remain uniformly overconfident (higher ECE) across all noise levels, while AUROC shows model-specific variance.}
\label{fig:calibration}
\vspace{-10pt}
\end{figure*}

\begin{table}[t]
\centering
\small
\caption{\small Clean-baseline calibration. Bold marks the better model per pair. ECE: Expected Calibration Error (lower is better). Mean Conf: average assigned confidence. AUROC: selective prediction AUROC (higher is better).}
\label{tab:rq2_calib_clean}
\setlength{\tabcolsep}{3pt}
\begin{tabularx}{\columnwidth}{l l C{0.9cm} C{1.0cm} C{0.9cm}}
\toprule
\textbf{Task} & \textbf{Model} & \textbf{ECE} & \textbf{Mean Conf} & \textbf{AUROC} \\
\midrule
\multirow{4}{*}{TriviaQA}
  & LLaDA-8B    & 0.814 & 0.999 & 0.617 \\
  & LLaMA-3-8B  & \cellcolor{blue!15}\textbf{0.292} & 0.837 & \textbf{0.627} \\
  \cmidrule(lr){2-5}
  & Dream-7B    & 0.740 & 0.990 & \cellcolor{blue!15}\textbf{0.707} \\
  & Qwen2.5-7B  & \cellcolor{blue!5}\textbf{0.535} & 0.852 & \cellcolor{blue!5}0.681 \\
\midrule
\multirow{4}{*}{GSM8K}
  & LLaDA-8B    & \textbf{0.351} & 0.976 & \cellcolor{blue!15}\textbf{0.681} \\
  & LLaMA-3-8B  & 0.362 & 0.821 & 0.527 \\
  \cmidrule(lr){2-5}
  & Dream-7B    & \cellcolor{blue!5}0.183 & 0.983 & 0.516 \\
  & Qwen2.5-7B  & \cellcolor{blue!15}\textbf{0.131} & 0.911 & \cellcolor{blue!5}\textbf{0.569} \\
\midrule
\multirow{4}{*}{ARC-Chall.}
  & LLaDA-8B    & 0.140 & 1.000 & \cellcolor{blue!15}\textbf{0.772} \\
  & LLaMA-3-8B  & \cellcolor{blue!15}\textbf{0.077} & 0.768 & 0.520 \\
  \cmidrule(lr){2-5}
  & Dream-7B    & 0.144 & 0.994 & \cellcolor{blue!5}\textbf{0.745} \\
  & Qwen2.5-7B  & \cellcolor{blue!5}\textbf{0.118} & 0.873 & 0.706 \\
\bottomrule
\end{tabularx}
\end{table}

\paragraph{Consistent DLM Overconfidence:}
\dlm{}s are systematically more confident than \ar models (Table~\ref{tab:rq2_calib_clean}, Figure~\ref{fig:baseline_crosspair}). For example, LLaDA averages near-perfect confidence ($0.976$--$1.000$), while LLaMA-3 scores much lower ($0.768$--$0.837$). This results in stark ECE gaps, particularly on TriviaQA ($+0.522$ and $+0.205$ for the two pairs). 

\paragraph{Confident Commitment Is a Hazard:}
Because \dlm{}s commit to a token precisely when its probability peaks, high marginal confidence is a mechanical byproduct of the architecture. However, this mechanism decouples confidence from correctness. Under heavy perturbation ($r{=}0.30$), \dlm confidence remains $\geq 0.93$ even as accuracy collapses, whereas \ar confidence appropriately drops alongside accuracy (Figure~\ref{fig:calibration}). In practical deployment, a model that remains highly confident while answering incorrectly is much harder to trust and filter than one that properly signals its uncertainty. The preserved AUROC scores (e.g., LLaDA GSM8K at $0.681$) confirm that the model retains discriminative signal; the issue lies in its absolute calibration.

\paragraph{Insensitivity of Per-Token Signals:}
This overconfidence also explains why internal token-level filters fail. \dlm{}s assign near-1.0 log-probabilities even to corrupted tokens (e.g., \texttt{teh}) because they evaluate each token within an otherwise clean, bidirectional context. \dlm robustness is a dynamic, global process that cannot be reliably localized to individual token probabilities.

\subsection{RQ3: Gradient Landscape via D-GCG Probe}
\label{sec:rq3}

\begin{figure}[t]
\centering
\includegraphics[width=\columnwidth]{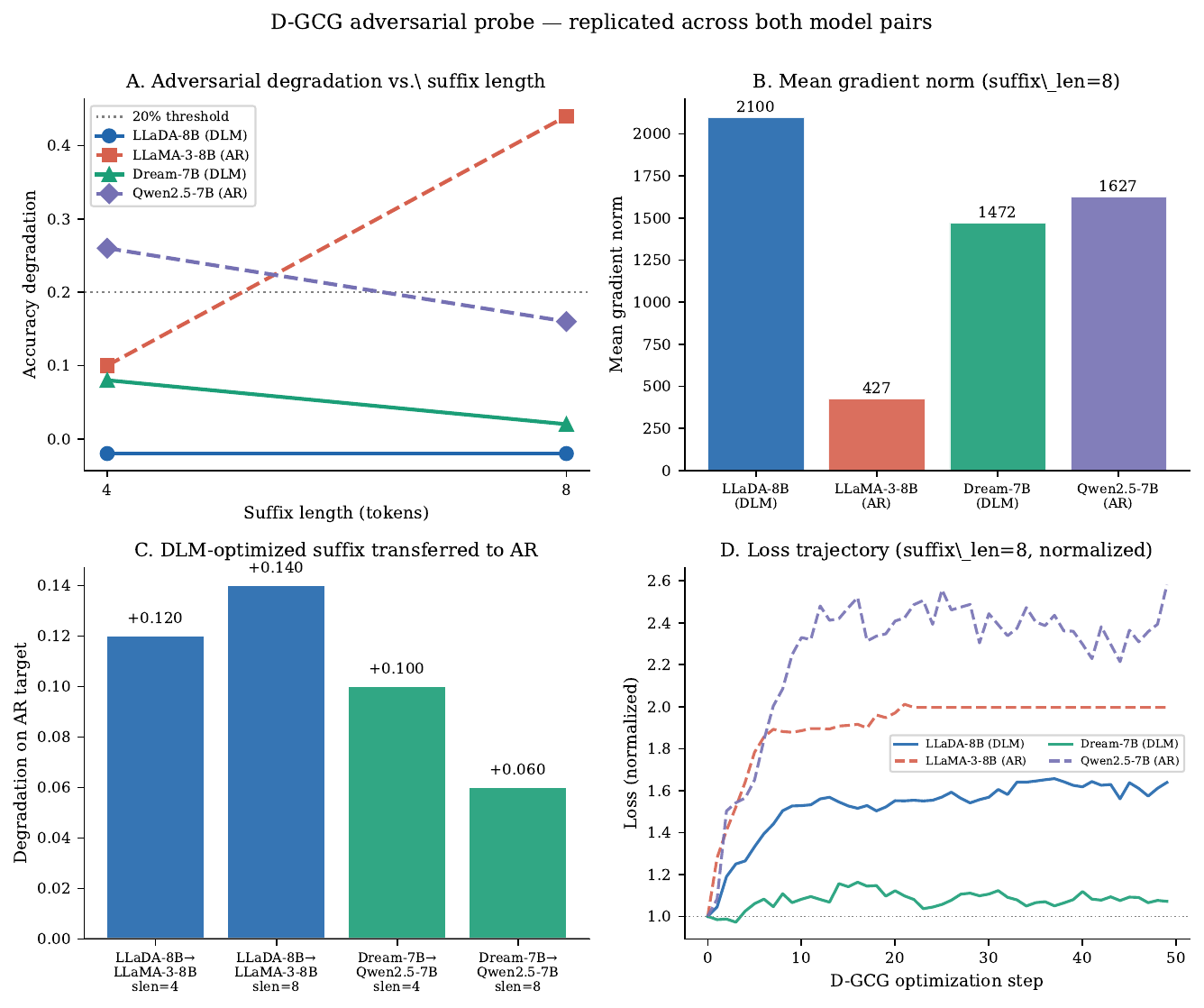}
\caption{\small D-GCG adversarial probe. \textbf{(A)} Accuracy degradation vs.\ suffix length: neither \dlm reaches the 20\% degradation threshold, while both \ar models exceed it at length $\leq 8$. \textbf{(B)} Mean gradient norms at length 8: \dlm gradients are comparable to (Dream) or much larger than (LLaDA) \ar gradients magnitude is not the issue. \textbf{(C)} Transferability to \ar targets is moderate. \textbf{(D)} Loss trajectory across 50 D-GCG optimization steps: \ar curves ascend consistently (successful attack), while \dlm curves oscillate erratically, showing the optimizer cannot exploit the \dlm gradient direction.}
\label{fig:dgcg}
\end{figure}

\begin{table}[t]
\centering
\small
\caption{\small D-GCG / AR-GCG probe results on TriviaQA. Min-len: shortest suffix causing $\geq$ 20\% accuracy drop. Best deg.: maximum degradation across suffix lengths 4 and 8. Grad norm: mean per-step gradient norm at length 8.}
\label{tab:rq3_dgcg}
\setlength{\tabcolsep}{3pt}
\begin{tabularx}{\columnwidth}{l C{1.2cm} C{1.3cm} C{1.2cm}}
\toprule
\textbf{Model} & \textbf{Min-len} & \textbf{Best deg.} & \textbf{Grad norm} \\
\midrule
LLaDA-8B (DLM)     & \textbf{None}  & $-0.020$ & 2{,}100 \\
LLaMA-3-8B (AR)    & 8              & $+0.440$ & 427    \\
\midrule
Dream-7B (DLM)     & \textbf{None}  & $+0.050$ & 1{,}472 \\
Qwen2.5-7B (AR)    & 4              & $+0.250$ & 1{,}627 \\
\bottomrule
\end{tabularx}
\end{table}

\paragraph{Adversarial Suffix Resistance:}
Under our short-suffix search budget (suffix lengths 4--8, 50 optimization steps, top-$k{=}64$, $n{=}50$), D-GCG fails to degrade \dlm performance. Neither \dlm reaches a 20\% degradation threshold; in fact, the optimizer marginally \emph{improves} LLaDA's accuracy ($-0.020$). In contrast, standard GCG easily breaks \ar models under the same budget, degrading LLaMA-3 by $+0.440$ and Qwen2.5 by $+0.250$. We therefore claim resistance to \emph{this class of short gradient-based suffix attacks}, not broad adversarial robustness; longer suffixes, larger budgets, and diffusion-native attacks~\citep{dija2026} remain untested.

\paragraph{Incoherent Gradients:}
This resistance is not due to vanishing gradients, \dlm gradient magnitudes actually match or exceed \ar gradients (Table~\ref{tab:rq3_dgcg}). Instead, the issue is gradient coherence. Figure~\ref{fig:dgcg}\,(D) illustrates this clearly. During optimization, \ar loss curves ascend steadily, indicating the attack successfully finds adversarial tokens. The \dlm loss curves, however, oscillate erratically. Despite large gradient magnitudes, the optimizer cannot successfully steer the \dlm.

\paragraph{Loss Landscape Dynamics:}
This incoherence stems directly from the diffusion training objective. The \dlm loss (Eq.~\ref{eq:loss}) averages over random timesteps and mask patterns. Consequently, a single-step gradient estimate is highly stochastic. The local gradient points in wildly different directions depending on which mask is sampled, making the effective loss surface too jagged for greedy coordinate substitution. Thus, \dlm resistance to GCG attacks is a natural byproduct of training-time noise marginalization, rather than an explicit defense mechanism.

\paragraph{Cross-Architecture Transferability:}
Suffixes optimized on \dlm{}s transfer to \ar targets with moderate success ($+0.12$ to $+0.15$ degradation), but they are notably weaker than native \ar attacks. This confirms that while the architectures share some representational overlap, their loss surfaces remain distinct.

\subsection{RQ4: Mechanistic Analysis}
\label{sec:rq4}

\begin{figure}[!tp]
\centering
\includegraphics[width=\columnwidth]{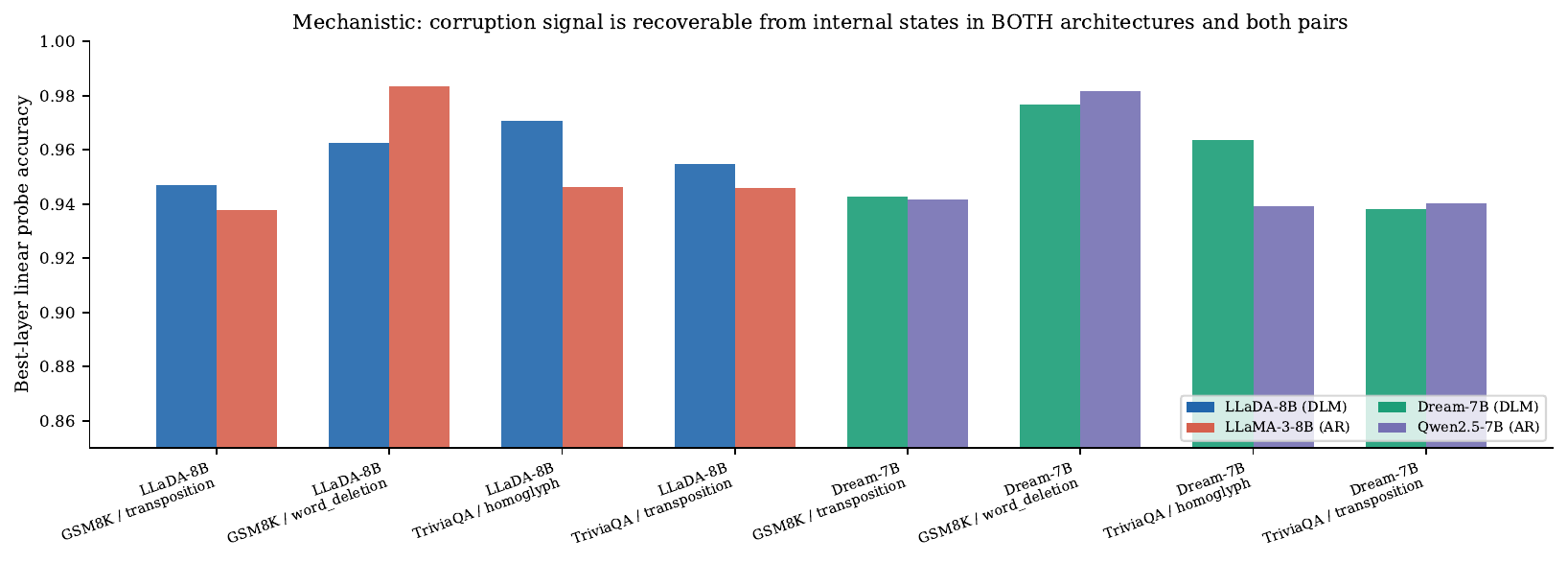}
\caption{\small Best-layer linear probe accuracy for detecting perturbed hidden states. All models exceed 93\% accuracy (best-layer AUROC 0.92--0.99, near-identical for \dlm and \ar), indicating corruption is linearly decodable in both architectures---consistent with a decoding-stage bottleneck rather than a failure to encode the corruption.}
\label{fig:mechanistic}
\end{figure}

We mechanistically probe 100 samples across four conditions per pair. Since \dlm{}s do not expose standard attention weights via their fused-kernel bidirectional implementations, our \dlm analysis relies on hidden-state probing and generation trajectories, whereas \ar baselines permit direct attention analysis.

\begin{table}[t]
\centering
\caption{\small Best-layer linear probe accuracy on clean-vs-noisy hidden state classification. Darker and lighter blue shading indicate the highest and second-highest probe accuracies per condition across the four models. All values exceed 93\% (best-layer AUROC 0.92--0.99), showing that corruption is linearly decodable in both architectures.}
\label{tab:mechanistic_crosspair}
\resizebox{\columnwidth}{!}{%
\begin{tabular}{l l c c c c}
\toprule
\multirow{2}{*}{\textbf{Pair}} & \multirow{2}{*}{\textbf{Model}} & \multicolumn{2}{c}{\textbf{GSM8K}} & \multicolumn{2}{c}{\textbf{TriviaQA}} \\
\cmidrule(lr){3-4} \cmidrule(lr){5-6}
 & & \textbf{Transposition} & \textbf{Word-deletion} & \textbf{Homoglyph} & \textbf{Transposition} \\
\midrule
\multirow{2}{*}{Pair 1}
  & LLaDA    & 0.947 & 0.963 & \cellcolor{blue!15}0.971 & \cellcolor{blue!5}0.955 \\
  & LLaMA-3   & 0.938 & \cellcolor{blue!5}0.984 & 0.946 & 0.946 \\
\midrule
\multirow{2}{*}{Pair 2}
  & Dream    & \cellcolor{blue!5}0.958 & 0.979 & 0.944 & \cellcolor{blue!15}0.964 \\
  & Qwen2.5   & \cellcolor{blue!15}0.969 & \cellcolor{blue!15}0.986 & \cellcolor{blue!5}0.953 & 0.954 \\
\bottomrule
\end{tabular}%
}
\end{table}

\paragraph{Corruption Is Linearly Encoded in Both Architectures:}
Hidden-state linear probes distinguish clean from perturbed inputs with $>0.93$ accuracy (best-layer AUROC 0.92--0.99) across all models (Table~\ref{tab:mechanistic_crosspair}, Figure~\ref{fig:mechanistic}). Crucially, the probe is near-identical for the \dlm and its \ar partner (e.g., transposition AUROC 0.982 vs.\ 0.983), so corruption is linearly decodable in \emph{both} architectures and cannot by itself explain their behavioral difference. Interestingly, while the corruption signal peaks in the middle-to-late layers for LLaDA, it peaks early for Dream and collapses by the final layer. Yet, in all cases, the corruption is present in states read by the decoder.

\begin{figure}[t]
\centering
\includegraphics[width=\columnwidth]{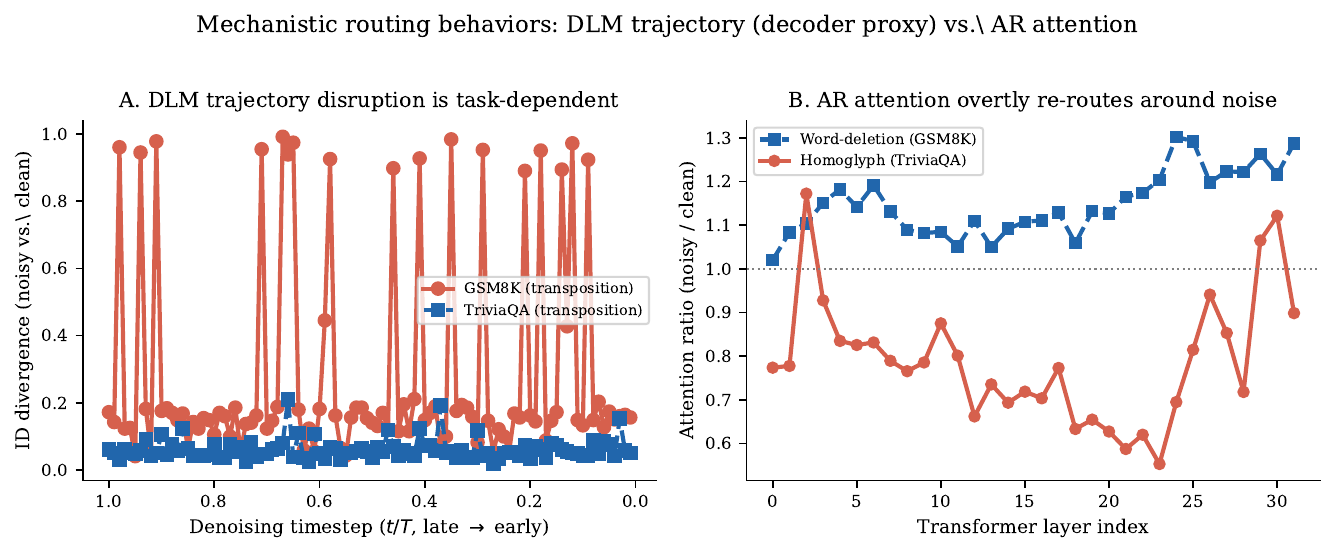}
\caption{\small Mechanistic routing behaviors. (A) DLM trajectory divergence varies significantly by task. (B) AR models exhibit active dynamic re-routing (e.g., suppressing homoglyphs or compensating for deletions), a compensatory mechanism absent in DLMs.}
\label{fig:routing}
\vspace{-10pt}
\end{figure}

\paragraph{Toward a Decoding-Stage Bottleneck:}
If the corruption is encoded, why do models fail? Figure~\ref{fig:routing} contrasts how each architecture reacts. \ar models actively shift their attention: for example, redistributing focus after word-deletion, or suppressing attention to homoglyphs. \dlm{}s, however, lack this overt compensatory routing. On GSM8K, LLaDA's denoising trajectory diverges by 28\% from its clean path, accompanied by a drop in confidence. Because the corruption is linearly decodable yet behavior still degrades, the fragility is \emph{consistent with} a decoding-stage bottleneck rather than an encoding failure. This probe is correlational; causally localizing the failure (e.g., via activation patching) is important future work.

\subsection{Pre-Generation Mitigation Fails (IPM)}
\label{sec:ipm}

Our finding that every model explicitly encodes corruption ($>0.93$ probe accuracy) leads to a clear hypothesis: pre-generation input patching should not fix \dlm fragility if the bottleneck lies at decoding rather than in the input the decoder receives. We test this using \textbf{Input Denoising via Partial Masking} (\ipm), which flags suspected corrupt tokens, masks them, and runs a short ($T_d{=}64$) diffusion infilling pass before standard generation.

\paragraph{Internal Detection Signal:}
We initially hypothesized that \dlm per-token conditional probabilities could serve as a corruption detector. However, as noted in Section~\ref{sec:rq2}, \dlm{}s assign near-1.0 log-probabilities even to corrupted tokens because they score them within an otherwise-clean context. As a result, internal-signal variants like \ipm-Threshold and \ipm-TopK perform worse than the noisy baseline.

\paragraph{Hybrid-IPM with External Signal:}
To bypass this, Hybrid-IPM uses an external spell-checker to flag out-of-vocabulary tokens (edit distance $\leq 2$) for re-masking. Yet, as Table~\ref{tab:ipm_variants} shows, this still degrades task accuracy (e.g., dropping GSM8K keyboard-adjacency from $0.495$ to $0.425$). The external dictionary mistakenly flags valid domain-specific terms (like math variables or entity names), forcing the \dlm to hallucinate replacements. Ultimately, none of the five tested \ipm variants outperform the noisy baseline. This is consistent with our mechanistic diagnosis: for these pre-generation patching variants, surface-level input edits do not improve \dlm robustness, pointing instead to a decoding-stage bottleneck.

\begin{table}[t]
\centering
\small
\caption{\small IPM variant accuracy on selected conditions. Darker / lighter blue mark best / second-best per column. None of the five tested \ipm variants improves over the noisy baseline, consistent with a decoding-stage bottleneck that pre-generation patching does not address.}
\label{tab:ipm_variants}
\setlength{\tabcolsep}{3pt}
\begin{tabularx}{\columnwidth}{l C{1.1cm} C{1.1cm} C{1.1cm} C{1.1cm}}
\toprule
\textbf{Variant} & \multicolumn{2}{c}{\textbf{GSM8K}} & \multicolumn{2}{c}{\textbf{TriviaQA}} \\
 & kbd & trans & kbd & trans \\
\midrule
Baseline (noisy)  & \cellcolor{blue!5}0.495 & \cellcolor{blue!15}0.350 & \cellcolor{blue!15}0.100 & 0.060 \\
\ipm-TopK (best)  & \cellcolor{blue!15}0.500 & 0.330 & 0.075 & 0.045 \\
\ipm-EditDist     & 0.475 & \cellcolor{blue!5}0.340 & \cellcolor{blue!5}0.090 & \cellcolor{blue!15}0.075 \\
\ipm-Hybrid       & 0.425 & 0.315 & 0.070 & \cellcolor{blue!5}0.070 \\
\bottomrule
\end{tabularx}
\end{table}

\section{Conclusion}
\label{sec:conclusion}

In this work, we present a systematic, two-pair evaluation of Diffusion Language Model (\dlm{}) robustness and calibration, effectively isolating architectural traits from model-specific artifacts. We find that robustness to natural input noise is weight-dependent rather than a guaranteed feature of the diffusion objective: LLaDA broadly outperforms LLaMA-3, but Dream's advantage over Qwen2.5 is task-specific. However, two architectural traits are universal: \dlm{}s exhibit systematic overconfidence---a practical deployment hazard---and they naturally resist gradient-based adversarial attacks due to their highly stochastic loss landscapes. Crucially, mechanistic probing reveals that all models perfectly encode input corruption, proving that behavioral fragility stems entirely from the bidirectional decoder failing to quarantine noisy tokens. Because this is a decoder failure, our tests confirm that surface-level prompt patching (\ipm{}) cannot resolve it. Ultimately, \dlm robustness must be fundamentally trained into the decoding process rather than patched on after the fact, a research direction we support by releasing all our code.
\section*{Limitations}
\label{sec:limitations}

\paragraph{Scale and Generality:} 
We evaluate models exclusively at the 7B--8B parameter scale. It remains an open question whether these robustness properties, calibration flaws, and routing behaviors shift at larger scales (e.g., 70B+), where emergent resilience often appears. Furthermore, because we demonstrate that natural robustness is weight-dependent, our findings are naturally tied to the specific training recipes of LLaDA and Dream. As diffusion pretraining techniques mature, future \dlm{}s may exhibit different behavioral profiles.

\paragraph{Task and Perturbation Scope:} 
While we test nine common real-world noise types across three factual and reasoning benchmarks, our taxonomy is not exhaustive. Future work should evaluate domain-specific corruptions (e.g., code syntax errors, multilingual switching) and longer-context tasks (such as document summarization or multi-turn dialogue) to verify if these trends hold in more complex environments.

\paragraph{Calibration Methodology:} 
We calculate \dlm{} confidence using the geometric mean of per-token softmax probabilities at the final committed step. While this is a principled metric derived directly from the generation mechanism, it is only one possible aggregation strategy. We explore alternative confidence metrics in Appendix~\ref{app:calib_ablation}.

\section*{Ethical Considerations}

This work evaluates the robustness of publicly released models on standard open-source benchmarks. We do not introduce new attack capabilities; our \dgcg{} method is strictly a diagnostic probe for analyzing loss landscapes, not an offensive jailbreak. We use no private or sensitive data, and our automated input perturbations do not generate harmful or toxic content.
\bibliography{references}

\clearpage
\newpage
\appendix

\section{Discussion}
\label{sec:discussion}

\paragraph{Weight-Dependent Robustness.}
Our results show that natural robustness is not an automatic feature of diffusion models. While LLaDA consistently outperforms LLaMA-3 by a wide margin, Dream shows no such advantage over Qwen2.5 on any of the three tasks (indistinguishable on TriviaQA/ARC, worse on GSM8K). This shifts the community's core question from \emph{``Are \dlm{}s inherently more robust?''} to \emph{``Which training recipes produce robust \dlm{}s?''} Ultimately, the burden of safety and reliability lies in a model's pretraining mix, not just its diffusion objective.

\paragraph{Replicated Architectural Traits.}
Despite differences in natural robustness, two traits replicate across both DLM architectures. First, \dlm{}s exhibit systematic overconfidence. This is a direct byproduct of the iterative-commitment decoder, which emits tokens precisely when their probability is highest. Second, \dlm{}s naturally resist \dgcg{} adversarial suffixes. Their highly stochastic loss landscapes yield large but directionally incoherent gradients, preventing coordinate-wise attacks from succeeding under our short-suffix search budget.

\paragraph{The Decoder Bottleneck.}
The most striking finding of this work is the consistently high linear-probe accuracy ($>0.93$) across all tested models. Because the corruption is linearly decodable in both architectures, behavioral fragility is consistent with a decoding-stage bottleneck rather than a perceptual failure. This mechanistic insight motivated a falsifiable prediction: surface-level prompt patching should not help. Our \ipm capstone (\S\ref{sec:ipm}) is consistent with this across five variants. If the field wants to improve \dlm robustness, defenses may need to intervene directly in the iterative decoding loop, not the input prompt.

\paragraph{Practical Implications.}
For real-world deployment, our findings suggest three immediate takeaways: (i) \dlm confidence scores require post-hoc calibration before being used for selective prediction; (ii) natural robustness must be audited on a per-model basis; and (iii) while \dlm{}s naturally resist suffix attacks, diffusion-specific jailbreaks~\citep{dija2026,priming2026,pad2025} remain an independent threat requiring distinct mitigations.


\section{IPM Algorithm (Full Pseudocode)}
\label{app:ipm_algo}

Algorithm~\ref{alg:ipm} provides the complete pseudocode for \ipm{}-Threshold. Other variants (\ipm{}-TopK, \ipm{}-Uniform, \ipm{}-Iterative) differ only in the mask selection criteria (line 7).

\begin{algorithm}[H]
\caption{Input Denoising via Partial Masking (\ipm{}-Threshold)}
\label{alg:ipm}
\begin{algorithmic}[1]
\Require Noisy prompt $\tilde{\mathbf{x}} \in \mathcal{V}^L$, DLM $p_\theta$, threshold $\tau$, denoising steps $T_d$, min-token-length $\ell_\text{min}$
\Ensure Denoised prompt $\hat{\mathbf{x}}$
\State \textit{// Phase 1: Noise Detection}
\State $B \leftarrow \tilde{\mathbf{x}}\mathbf{1}^\top$ \Comment{$L$ copies of $\tilde{\mathbf{x}}$, shape $(L, L)$}
\For{$i = 1, \ldots, L$}
    \State $B[i, i] \leftarrow \textsc{[mask]}$ \Comment{Mask position $i$}
\EndFor
\State $\mathbf{s} \leftarrow \texttt{batch\_forward}(B)$ \Comment{$(L,)$ log-probs via chunks of 16}
\State \textit{// Phase 2: Targeted Re-masking}
\State $\mathcal{C} \leftarrow \{i : s_i < \log\tau \;\wedge\; \text{len}(\tilde{x}_i) \geq \ell_\text{min}\}$
\State $\hat{\mathbf{x}} \leftarrow \tilde{\mathbf{x}}$
\For{$i \in \mathcal{C}$}
    \State $\hat{x}_i \leftarrow \textsc{[mask]}$
\EndFor
\State \textit{// Phase 3: Short Diffusion Fill-in}
\For{$t = T_d, T_d - 1, \ldots, 1$}
    \State Compute logits $\leftarrow p_\theta(\hat{\mathbf{x}})$
    \State Unmask $\lceil(T_d - t + 1)/T_d \cdot |\mathcal{C}|\rceil$ most-confident positions
\EndFor
\State \Return $\hat{\mathbf{x}}$
\end{algorithmic}
\end{algorithm}

\paragraph{Complexity.}
Phase 1 requires $\lceil L/16 \rceil$ batched forward passes (batch size 16, sequence length $L$). Phase 3 requires $T_d$ forward passes on a single sequence. The total overhead is roughly $L/16 + T_d$ forward passes. For standard generation parameters ($L \approx 64$, $T_d = 16$, $T = 128$), the overhead is just 20 additional passes, yielding $\sim 1.5\times$ the base inference latency.

\section{D-GCG Implementation Details}
\label{app:dgcg_details}

\paragraph{Gradient Computation.}
To compute gradients with respect to discrete suffix tokens, we use a straight-through differentiable embedding. The suffix is represented as a one-hot matrix $\mathbf{E} \in \{0,1\}^{K \times |\mathcal{V}|}$ and passed through the embedding matrix $\mathbf{W}_e$ as $\mathbf{S} = \mathbf{E}\mathbf{W}_e \in \R^{K \times d}$. Gradients $\partial \mathcal{L} / \partial \mathbf{E}$ are computed via standard backpropagation; the argmax of this gradient over the vocabulary identifies the top-$k$ candidate tokens~\citep{zou2023universal}.

\paragraph{Hyperparameters.}
Our search space includes suffix lengths $K \in \{4, 8, 16\}$, 50 optimization steps, 64 top-$k$ candidates, and a mini-batch size of $1$ per gradient step, evaluated on $n{=}50$ samples. We sample $n_\text{timestep}=1$ (single-timestep estimation, see \S\ref{sec:dgcg}). The suffix is initialized randomly from the vocabulary. For cross-pair comparability we report results up to length 8 (not all configurations completed length 16), matching the main-text figures and Table~\ref{tab:rq3_dgcg}.

\paragraph{VRAM Budget.}
The single-step gradient estimator requires approximately $1.5\times$ standard inference VRAM ($\approx 8$\,GB on an RTX 4090). We do not perform backpropagation through time (BPTT) across the $T$ denoising steps.

\section{Prompt Templates}
\label{app:prompts}

Models receive prompts formatted through their official chat templates via \texttt{tokenizer.apply\_chat\_template}. The user message for each benchmark is constructed as follows:

\paragraph{TriviaQA.}
\begin{verbatim}
Answer the following question concisely
in a few words.

Question: {question}
Answer:
\end{verbatim}

\paragraph{GSM8K.}
\begin{verbatim}
Solve the following math problem step
by step. At the end, write the final
answer as a number after '####'.

Problem: {question}
Solution:
\end{verbatim}

\paragraph{ARC-Challenge.}
\begin{verbatim}
Answer the following multiple choice
science question. Reply with just the
letter (A, B, C, or D).

Question: {question}
Choices:
  A) {choice_A}
  B) {choice_B}
  C) {choice_C}
  D) {choice_D}
Answer:
\end{verbatim}

\section{Full Hyperparameter Tables}
\label{app:hyperparams}

This section details the complete set of hyperparameters used across all experiments. Table~\ref{tab:hyperparams_model} outlines the generation and architectural settings for both the \dlm{} and \ar models. Table~\ref{tab:hyperparams_ipm} specifies the grid of parameters tested during our \ipm{} ablation study, and Table~\ref{tab:hyperparams_calib} details the configurations used for our calibration metrics.

\begin{table}[htbp]
\centering
\small
\caption{Model hyperparameters used in all experiments.}
\label{tab:hyperparams_model}
\resizebox{\columnwidth}{!}{
\begin{tabular}{lcc}
\toprule
\textbf{Parameter} & \textbf{DLMs} & \textbf{AR models} \\
\midrule
Quantization & none (BF16) & none (BF16) \\
Compute dtype & bfloat16 & bfloat16 \\
Device map & auto & auto \\
Diffusion steps ($T$) & 128 (LLaDA) / 512 (Dream) & --- \\
LLaDA block length & 32 & --- \\
LLaDA EOS suppression & on (App.~B.4 trick) & --- \\
Dream remasking alg.\ & \texttt{entropy} ($\alpha_T{=}0$) & --- \\
Dream temperature & 0.2 & --- \\
Dream top-$p$ & 0.95 & --- \\
Max new tokens & 128 (512 for GSM8K) & 128 (512 for GSM8K) \\
Temperature (default) & 0 (greedy) & 0 (greedy) \\
Attention impl. & eager & sdpa \\
\bottomrule
\end{tabular}
}
\end{table}

\begin{table}[htbp]
\centering
\small
\caption{\ipm{} hyperparameters evaluated in the ablation study.}
\label{tab:hyperparams_ipm}
\resizebox{\columnwidth}{!}{
\begin{tabular}{ll}
\toprule
\textbf{Parameter} & \textbf{Values} \\
\midrule
Variant & Threshold, TopK, Uniform, Iterative \\
Threshold $\tau$ & 0.05, 0.10, 0.20, 0.40 \\
TopK $K$ & 1, 2, 4, 8 \\
Mask fraction $\rho$ & 0.05, 0.10, 0.20 \\
Denoising steps $T_d$ & 16 \\
Min token length $\ell_\text{min}$ & 3 characters \\
Max iterations (Iterative) & 3 \\
\bottomrule
\end{tabular}
}
\end{table}

\begin{table}[htbp]
\centering
\small
\caption{Calibration computation details.}
\label{tab:hyperparams_calib}
\resizebox{\columnwidth}{!}{
\begin{tabular}{ll}
\toprule
\textbf{Parameter} & \textbf{Value} \\
\midrule
ECE bins & 15 (equal-width) \\
DLM confidence aggregation & Geometric mean \\
AUROC estimator & Trapezoidal (scikit-learn) \\
ELBO samples for scoring & 10 \\
\bottomrule
\end{tabular}
}
\end{table}

\paragraph{DLM Sampler Configurations (Reproducibility Note).}
Both \dlm{}s are evaluated using their authors' officially recommended generation recipes, which differ significantly from Hugging Face defaults:
\begin{itemize}
    \item \textbf{LLaDA-8B:} Uses semi-autoregressive block decoding with \texttt{block\_length}=32, low-confidence remasking, $T{=}128$, and EOS/EoT suppression in the confidence ranking (LLaDA paper App.\ B.4).
    \item \textbf{Dream-7B:} Uses \texttt{alg}=``entropy'' confidence remasking with $T{=}512$ (committing one token per step), \texttt{alg\_temp}=$0$, temperature $0.2$, and top-$p$ $0.95$. \emph{Note: The Hugging Face default for Dream (\texttt{alg}=``origin'', purely random unmasking) degrades long-output tasks by 3--4$\times$ and should not be used.}
\end{itemize}
Additionally, pipeline components that read DLM logits directly (\ipm{} noise detection, infilling, \dgcg{} gradient computation) apply Dream's causal-shift correction (\texttt{logits[:,\,:-1,\,:]}) to ensure token $i$ is scored from the logit at position $i{-}1$.

\section{Full RDI Curves by Perturbation Type}
\label{app:full_results}

\subsection*{Both Pairs, Side-by-Side (TriviaQA and ARC-Challenge)}

\begin{figure*}[htbp]
\centering
\includegraphics[width=\textwidth]{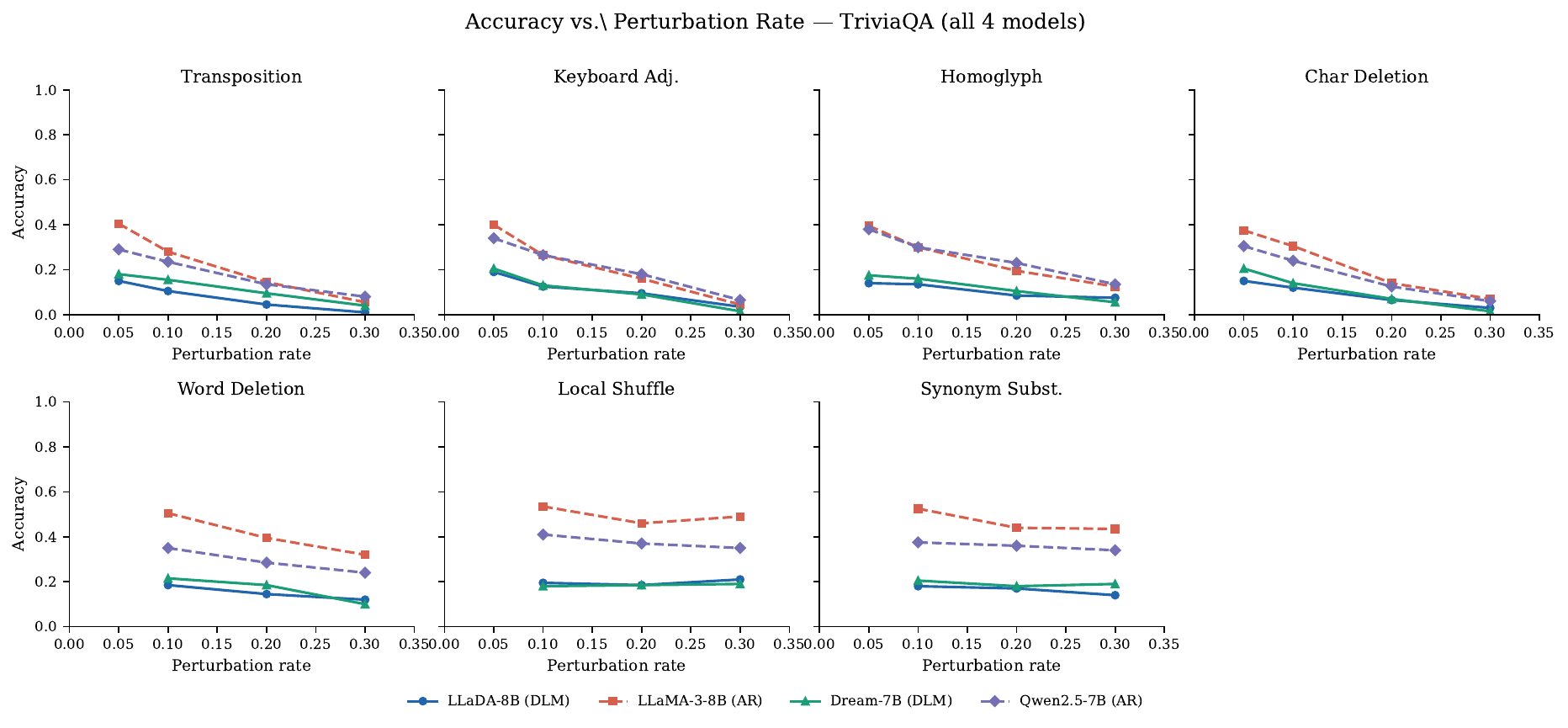}
\caption{Accuracy vs.\ perturbation rate on TriviaQA. \textbf{Top:} LLaDA-8B vs.\ LLaMA-3-8B. \textbf{Bottom:} Dream-7B vs.\ Qwen2.5-7B.}
\label{fig:sweep_triviaqa}\label{app:dream_sweep_triviaqa}
\end{figure*}

\begin{figure*}[htbp]
\centering
\includegraphics[width=\textwidth]{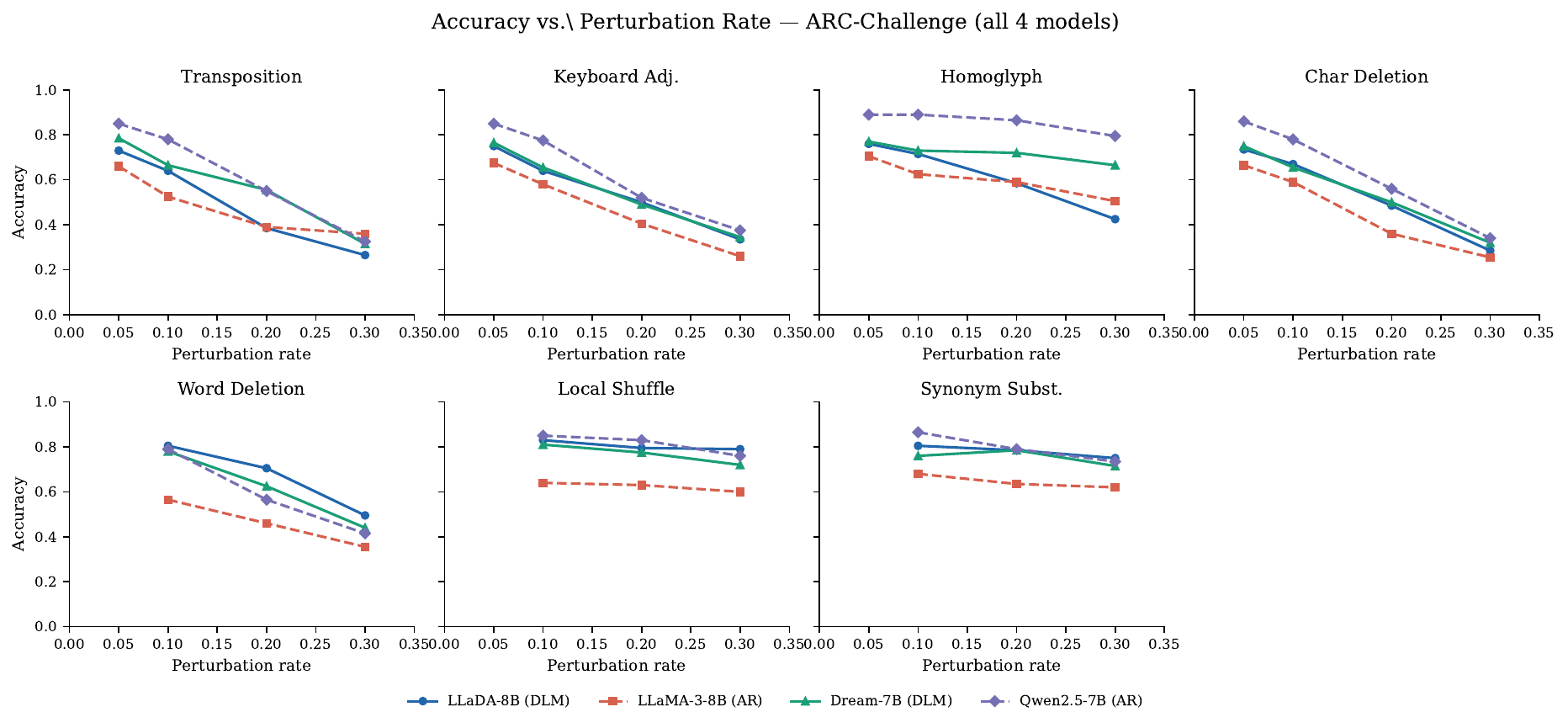}
\caption{Accuracy vs.\ perturbation rate on ARC-Challenge. \textbf{Top:} LLaDA pair. \textbf{Bottom:} Dream pair.}
\label{fig:sweep_arc}\label{app:dream_sweep_arc}
\end{figure*}

\begin{figure*}[htbp]
\centering
\includegraphics[width=\textwidth]{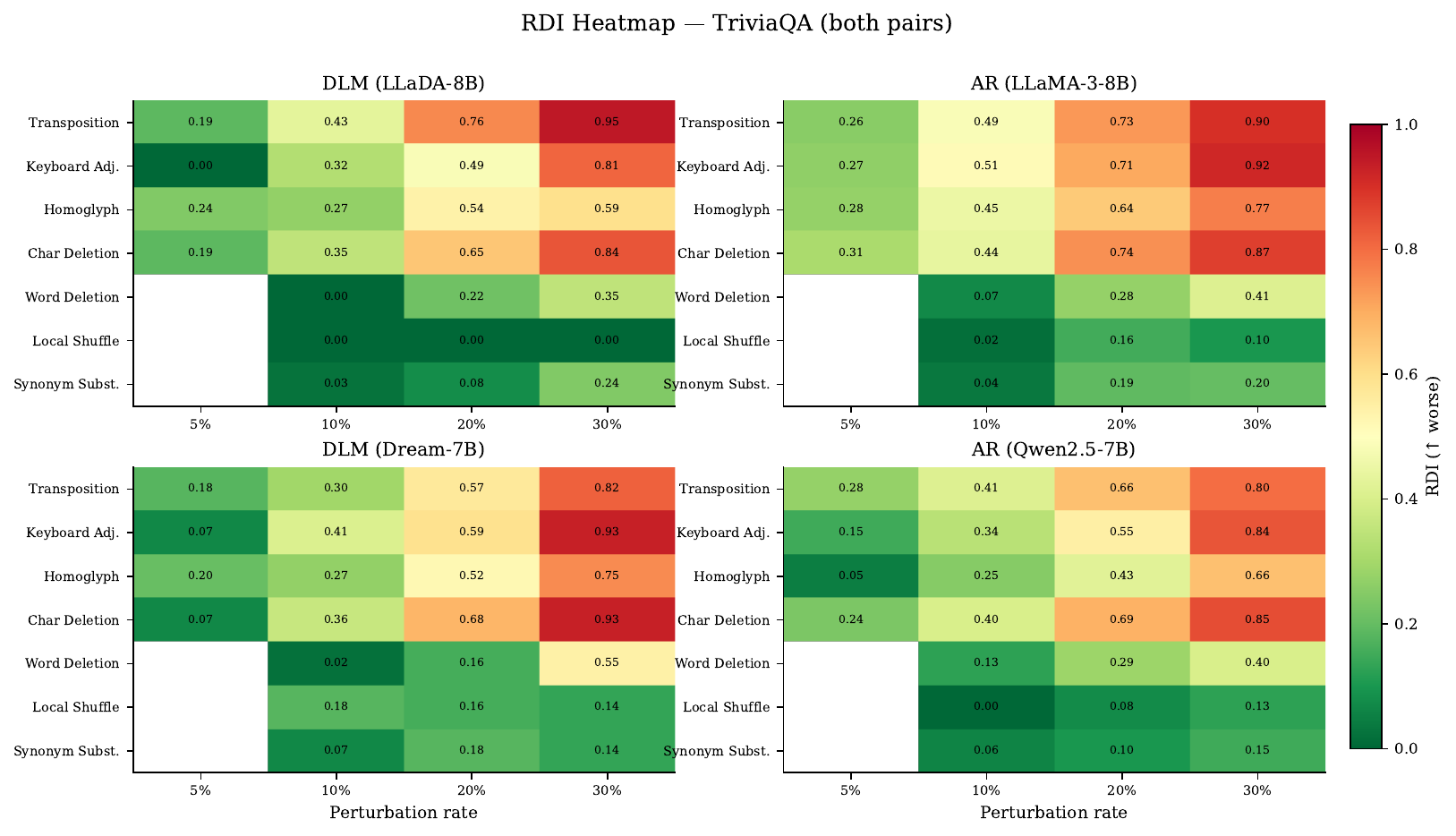}
\caption{RDI heatmap on TriviaQA for both pairs.}
\label{fig:rdi_triviaqa}\label{app:dream_rdi_triviaqa}
\end{figure*}

\begin{figure*}[htbp]
\centering
\includegraphics[width=\textwidth]{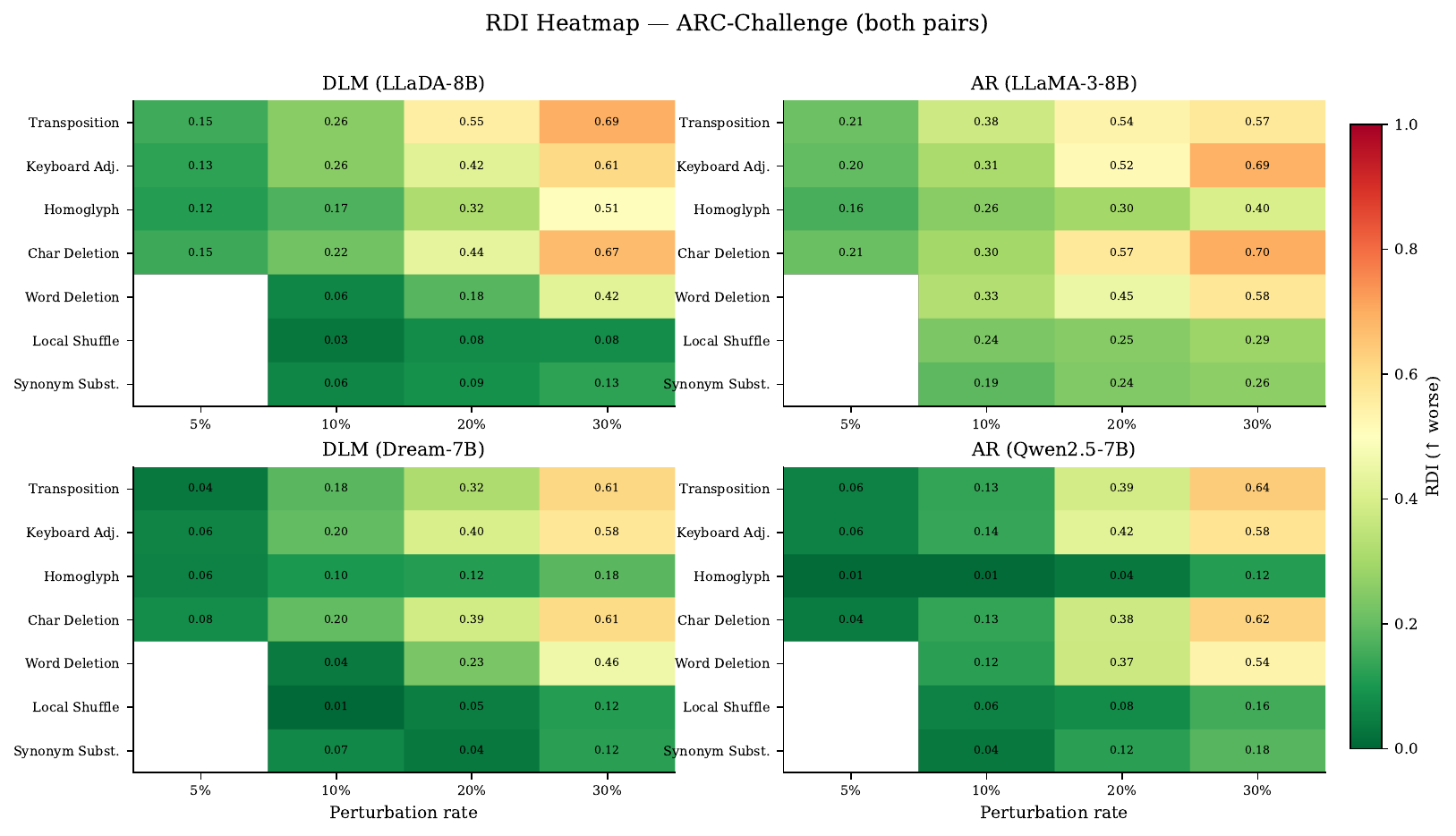}
\caption{RDI heatmap on ARC-Challenge for both pairs.}
\label{fig:rdi_arc}\label{app:dream_rdi_arc}
\end{figure*}

\begin{figure*}[htbp]
\centering
\includegraphics[width=\textwidth]{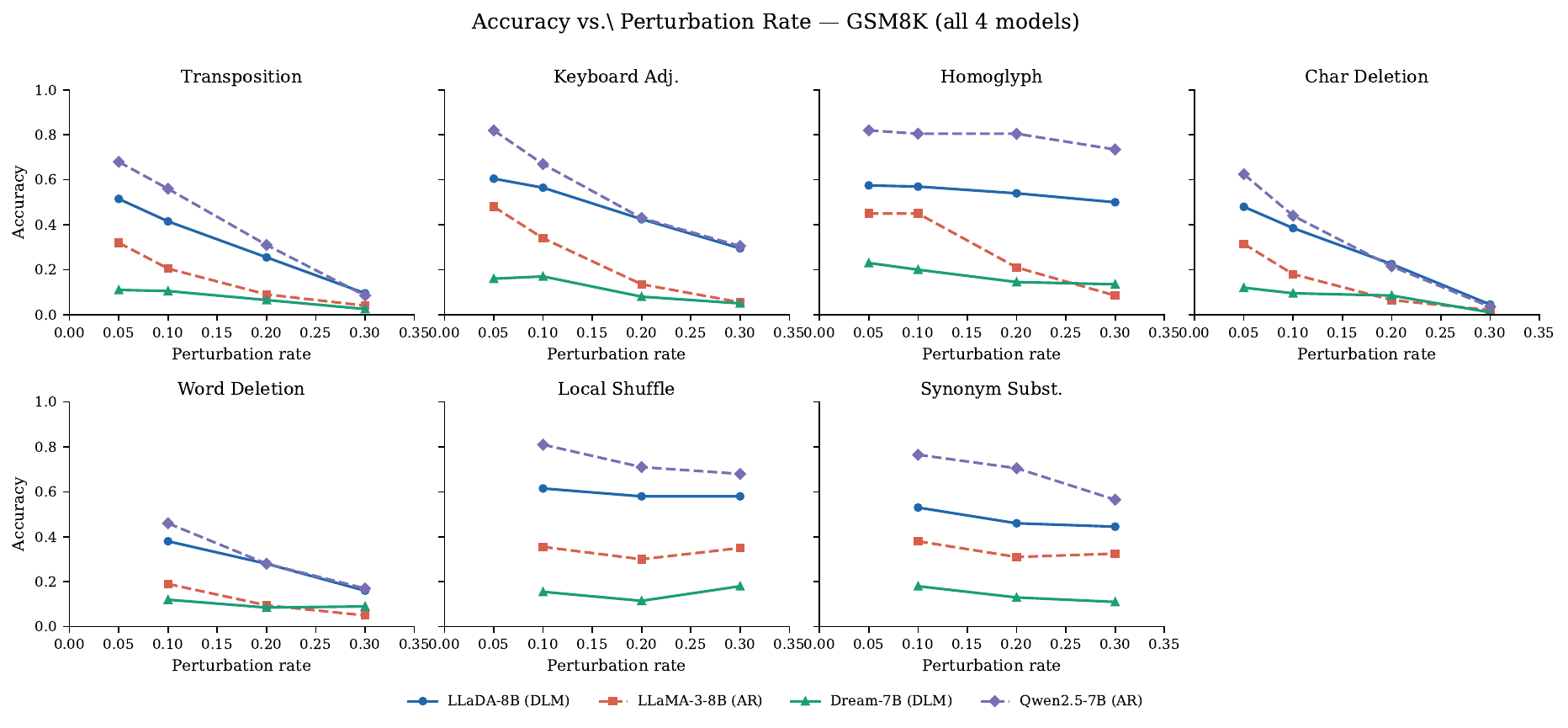}
\caption{Accuracy vs.\ perturbation rate on GSM8K (reproduced from Fig.~\ref{fig:sweep_gsm8k} for completeness).}
\label{app:dream_sweep_gsm8k}
\end{figure*}

\begin{figure*}[htbp]
\centering
\includegraphics[width=\textwidth]{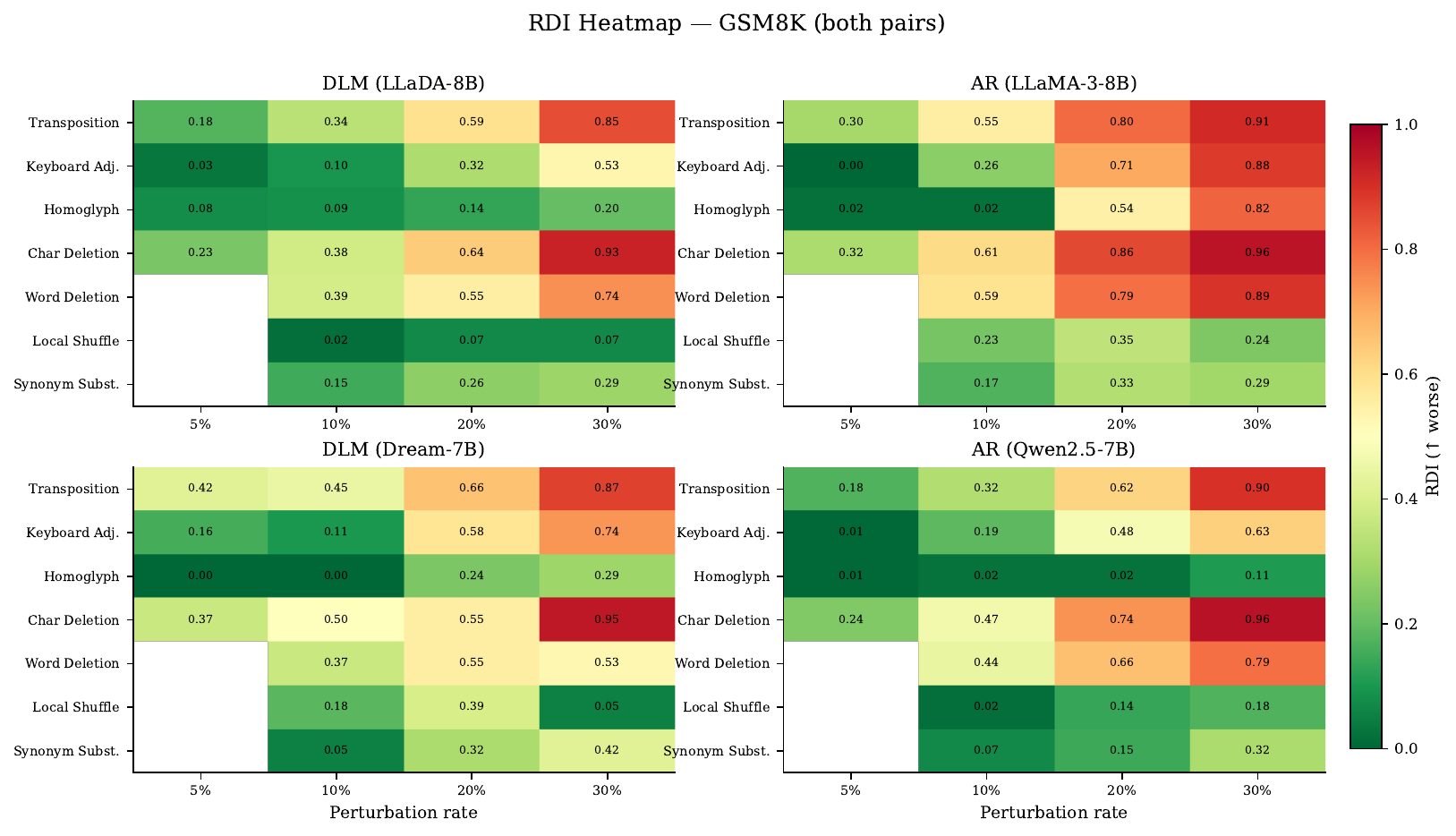}
\caption{RDI heatmap on GSM8K. Provides a complementary view to Figure~\ref{fig:sweep_gsm8k}, where cell color summarizes the degradation at specific perturbation rates.}
\label{app:dream_rdi_gsm8k}
\end{figure*}

To provide a complete view of model degradation under noise, we present the full accuracy sweep curves and corresponding RDI heatmaps for all tested benchmarks. Figures~\ref{fig:sweep_triviaqa} and \ref{fig:sweep_arc} plot the absolute accuracy degradation across varying perturbation rates for TriviaQA and ARC-Challenge, respectively, while Figure~\ref{app:dream_sweep_gsm8k} reproduces the GSM8K curves for completeness. The corresponding RDI heatmaps—visualizing the relative robustness degradation for each specific noise type—are shown in Figures~\ref{fig:rdi_triviaqa}, \ref{fig:rdi_arc}, and \ref{app:dream_rdi_gsm8k}.

\subsection*{Per-Pair Summary Panels}

\begin{figure}[htbp]
\centering
\includegraphics[width=\columnwidth]{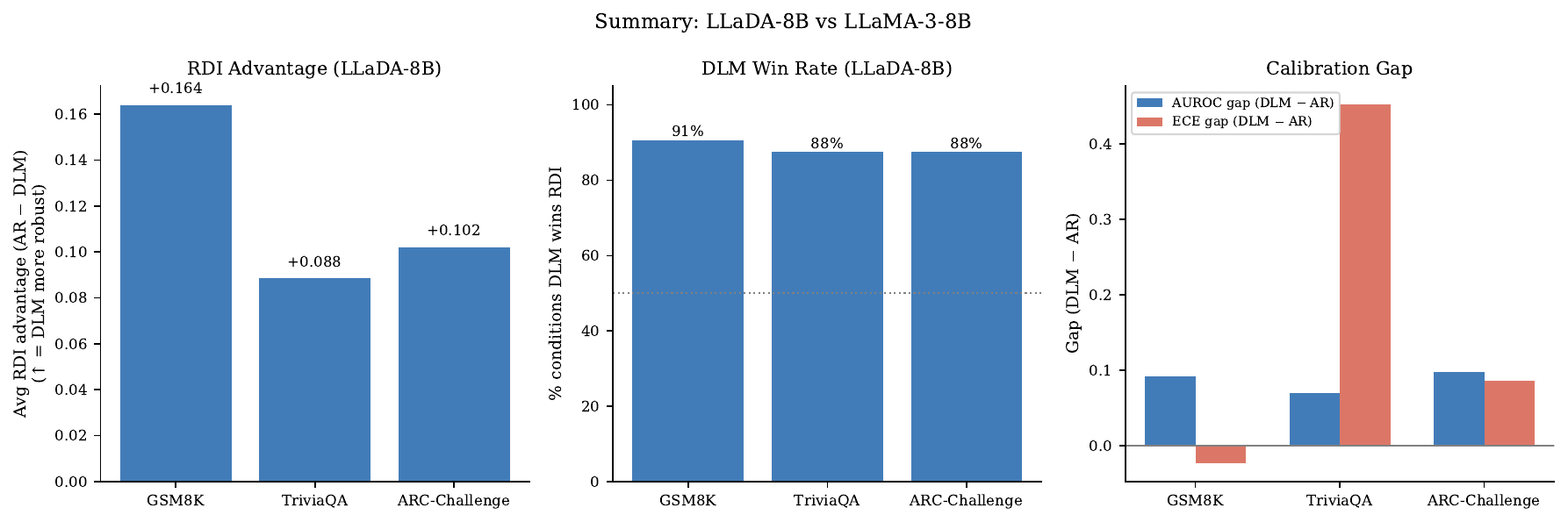}
\caption{LLaDA pair summary: RDI advantage, win rate, and calibration gap.}
\label{fig:summary_llada}
\end{figure}

\begin{figure}[htbp]
\centering
\includegraphics[width=\columnwidth]{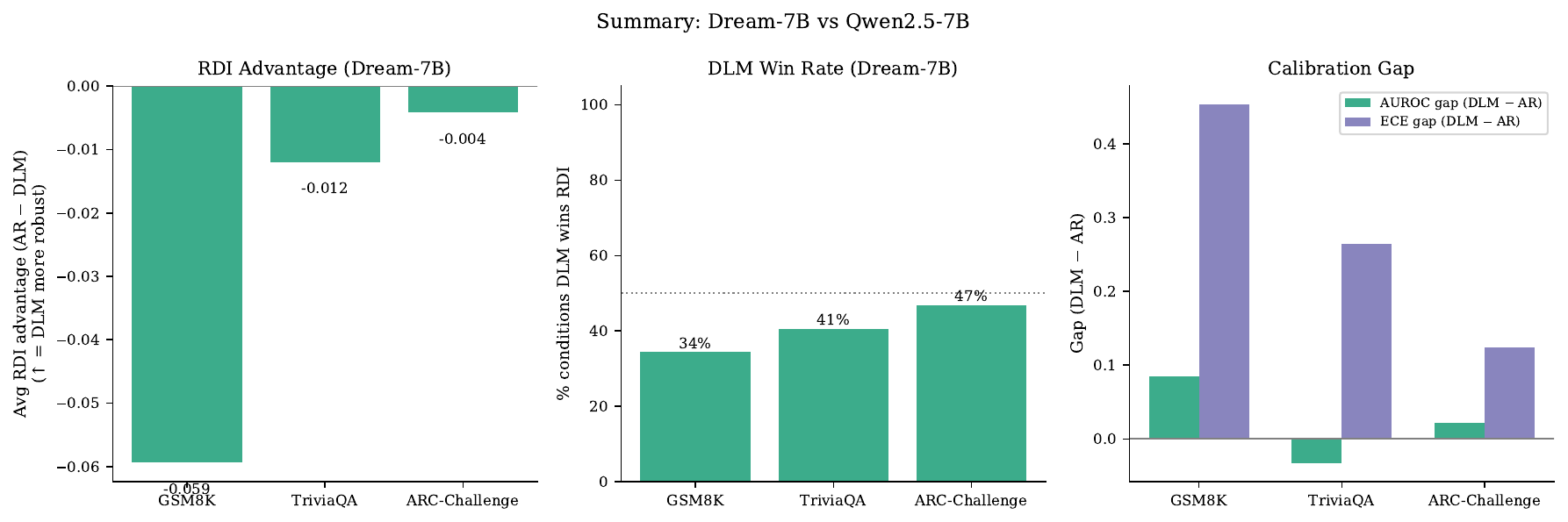}
\caption{Dream pair summary: RDI advantage (negative or near-zero, contrasting with the LLaDA pair), win rate (below 50\% on all three benchmarks), and calibration gap (demonstrating replicated DLM overconfidence).}
\label{app:summary_dream}
\end{figure}

To easily contrast the overall behavioral profiles of the two architectures, Figures~\ref{fig:summary_llada} and \ref{app:summary_dream} summarize the RDI advantage, win rates, and calibration gaps for the LLaDA and Dream pairs, respectively. Additionally, Table~\ref{tab:rq1_bytype} breaks down the Area Under the RDI curve (AURDI) by specific perturbation types for the LLaDA pair, providing granular insight into which noise categories drive the observed architectural differences.

\begin{table}[htbp]
\centering
\small
\caption{AURDI by perturbation type for the LLaDA pair on TriviaQA and GSM8K ($\downarrow$ = lower is more robust). The DLM yields a better (lower) AURDI across every perturbation type, detailing the win-rate results in \S\ref{sec:rq1}.}
\label{tab:rq1_bytype}
\resizebox{\columnwidth}{!}{
\begin{tabular}{ll cc cc}
\toprule
& & \multicolumn{2}{c}{\textbf{TriviaQA}} & \multicolumn{2}{c}{\textbf{GSM8K}} \\
\cmidrule(lr){3-4} \cmidrule(lr){5-6}
\textbf{Level} & \textbf{Type} & LLaDA & LLaMA-3 & LLaDA & LLaMA-3 \\
\midrule
\multirow{5}{*}{Char}
  & Transposition & 0.641 & 0.645 & 0.525 & 0.701 \\
  & Deletion      & 0.551 & 0.635 & 0.580 & 0.749 \\
  & Insertion     & 0.470 & 0.584 & 0.348 & 0.551 \\
  & Keyboard      & 0.454 & 0.647 & 0.266 & 0.537 \\
  & Homoglyph     & 0.441 & 0.573 & 0.129 & 0.389 \\
\midrule
\multirow{4}{*}{Word}
  & Deletion          & 0.196 & 0.259 & 0.560 & 0.766 \\
  & Local shuffle     & 0.000 & 0.108 & 0.058 & 0.291 \\
  & Synonym sub.      & 0.108 & 0.156 & 0.242 & 0.280 \\
  & Repetition        & 0.007 & 0.037 & 0.106 & 0.201 \\
\bottomrule
\end{tabular}
}
\end{table}

\paragraph{Observations.}
Homoglyph substitution is consistently the most damaging character-level perturbation for \dlm{}s, but not for \ar models. We hypothesize this is because LLaDA's tokenizer (derived from LLaMA-3) maps Cyrillic homoglyphs to out-of-vocabulary or rare tokens, producing unusual embedding vectors that confuse bidirectional attention. Furthermore, word-level perturbations are disproportionately damaging to \dlm{}s across both tasks.

\section{Full IPM Ablation Results}
\label{app:ipm_full}\label{app:ipm_ablation}

\begin{figure*}[htbp]
\centering
\includegraphics[width=\textwidth]{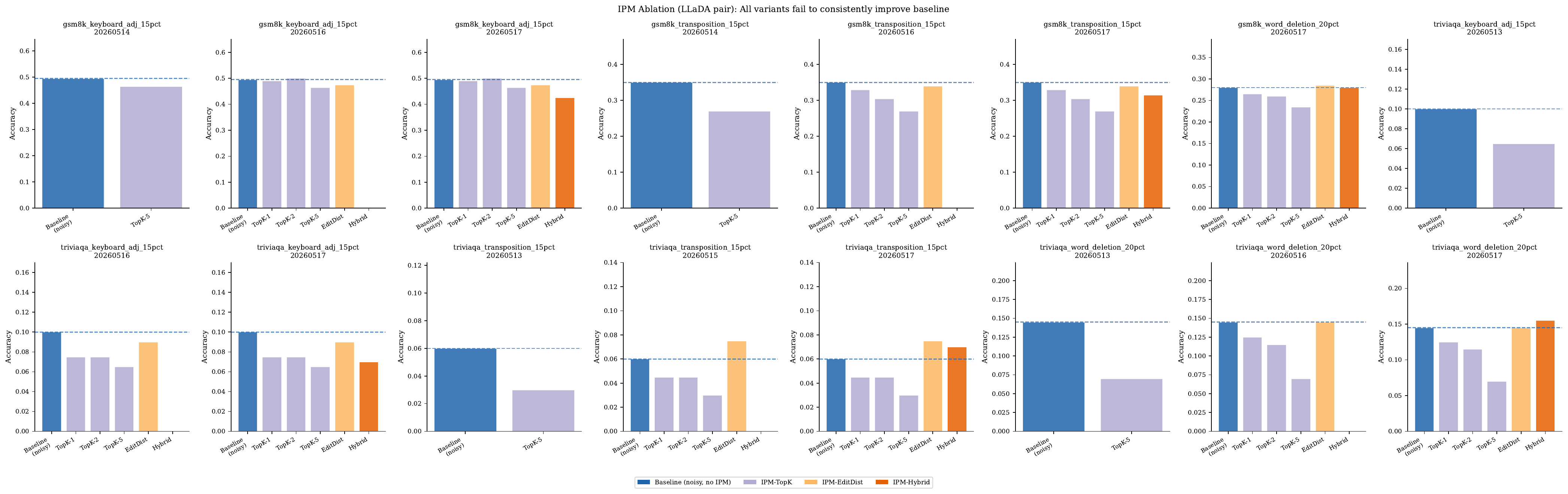}
\caption{\ipm{} variant comparison across all evaluated conditions. The dashed line represents the noisy baseline. No variant reliably improves over the baseline, and Hybrid-IPM (orange) consistently degrades accuracy.}
\label{fig:ipm_ablation}
\end{figure*}

The full ablation reported here was conducted on the \textbf{LLaDA pair} ($N{=}200$ samples), as LLaDA's generation recipe remained constant. The Dream-pair \ipm{} numbers discussed in the main text (Table~\ref{tab:ipm_variants}) stem from a subsequent $N{=}20$ run using Dream's corrected sampler. Figure~\ref{fig:ipm_ablation} visualizes the performance of all tested \ipm{} variants against the noisy baseline across multiple conditions. Furthermore, Table~\ref{tab:ipm_gsm8k} provides a detailed numerical breakdown of the ablation on GSM8K under keyboard-adjacency noise. 

Crucially, the qualitative pattern replicates across both models: no \ipm{} variant reliably improves over the noisy baseline. Aggressive Threshold variants tend to over-mask valid tokens, collapsing accuracy to chance (e.g., Dream TriviaQA-transposition Threshold-30 drops from $0.150$ to $0.000$). Small, single-sample upticks for Hybrid and Edit-Distance variants ($\pm 1$/$20$ samples) fall well within standard sampling noise and do not alter our conclusion that pre-generation patching is ineffective.

\begin{table}[htbp]
\centering
\small
\caption{Full \ipm{} ablation on GSM8K under keyboard-adjacency noise ($r{=}0.15$) for the LLaDA pair. Clean DLM accuracy is 0.625. No variant reliably improves over the noisy baseline; aggressive variants collapse accuracy.}
\label{tab:ipm_gsm8k}
\resizebox{\columnwidth}{!}{
\begin{tabular}{l ccc}
\toprule
\textbf{Method} & \textbf{Acc} & \textbf{$\Delta$Acc} & \textbf{Mean tokens changed} \\
\midrule
Baseline (noisy)                & 0.495 & ---      & --- \\
\midrule
\ipm-TopK ($K{=}1$)               & 0.490 & $-0.005$ & 0.5 \\
\ipm-TopK ($K{=}2$)               & 0.500 & $+0.005$ & 1.0 \\
\ipm-TopK ($K{=}3$)               & 0.500 & $+0.005$ & 1.5 \\
\ipm-TopK ($K{=}5$)               & 0.465 & $-0.030$ & 2.4 \\
\ipm-TopK ($K{=}10$)              & 0.480 & $-0.015$ & 4.6 \\
\midrule
\ipm-Uniform ($\rho{=}0.10$)      & 0.440 & $-0.055$ & 3.2 \\
\midrule
\ipm-Threshold ($\tau{=}0.10$)    & 0.395 & $-0.100$ & 17.5 \\
\ipm-Threshold ($\tau{=}0.20$)    & 0.325 & $-0.170$ & 19.7 \\
\ipm-Threshold ($\tau{=}0.30$)    & 0.325 & $-0.170$ & 20.9 \\
\ipm-Threshold ($\tau{=}0.50$)    & 0.285 & $-0.210$ & 23.0 \\
\midrule
\ipm-Iterative ($\times 3$)       & 0.300 & $-0.195$ & 24.0 \\
\ipm-EditDistance (external)      & 0.475 & $-0.020$ & 8.2 \\
\ipm-Hybrid (spell-checker)       & 0.425 & $-0.070$ & 9.0 \\
\bottomrule
\end{tabular}
}
\end{table}

\section{Statistical Significance and Question-Only Ablation}
\label{app:significance}

\paragraph{Paired significance of the RDI advantage.}
We treat the 32 noise conditions as paired observations of the RDI advantage (AR $-$ DLM) and report 95\% bootstrap confidence intervals (10k resamples) and Wilcoxon signed-rank tests (Table~\ref{tab:significance}). The LLaDA-pair advantage is significant and positive on all three tasks; the Dream-pair advantage is indistinguishable from zero on TriviaQA and ARC and significantly \emph{negative} on GSM8K. This places the ``robustness does not replicate across pairs'' conclusion on firm statistical footing.

\begin{table}[htbp]
\centering
\small
\caption{Paired bootstrap 95\% CI and Wilcoxon signed-rank test for the mean RDI advantage (AR $-$ DLM) over 32 conditions ($n{=}200$ each).}
\label{tab:significance}
\resizebox{\columnwidth}{!}{%
\begin{tabular}{ll cc c}
\toprule
\textbf{Pair} & \textbf{Task} & \textbf{Adv.} & \textbf{95\% CI} & \textbf{Wilcoxon $p$} \\
\midrule
\multirow{3}{*}{LLaDA} & TriviaQA & $+0.088$ & $[+0.060,+0.118]$ & $<0.001$ \\
                       & GSM8K    & $+0.164$ & $[+0.116,+0.216]$ & $<0.001$ \\
                       & ARC      & $+0.102$ & $[+0.070,+0.133]$ & $<0.001$ \\
\midrule
\multirow{3}{*}{Dream} & TriviaQA & $-0.012$ & $[-0.044,+0.020]$ & $0.488$ \\
                       & GSM8K    & $-0.059$ & $[-0.102,-0.016]$ & $0.013$ \\
                       & ARC      & $-0.004$ & $[-0.025,+0.017]$ & $0.665$ \\
\bottomrule
\end{tabular}%
}
\end{table}

\paragraph{Question-only perturbation.}
Our main sweep perturbs the entire formatted prompt. To confirm the cross-pair pattern is not an artifact of corrupting instruction/format tokens, we re-ran a reduced sweep perturbing \emph{only the raw question span} (mean over transposition/homoglyph/word-deletion at $r{=}0.20$; $n{=}100$ LLaDA pair, $n{=}50$ Dream pair). Table~\ref{tab:qonly} shows the LLaDA-pair DLM advantage persists (mean $+0.119$) while the Dream-pair effect stays near the noise floor (mean $+0.063$), matching the whole-prompt asymmetry.

\begin{table}[htbp]
\centering
\small
\caption{Question-only perturbation ablation: mean RDI over three conditions at $r{=}0.20$. Adv.\ $=$ AR RDI $-$ DLM RDI (positive $\Rightarrow$ DLM more robust).}
\label{tab:qonly}
\resizebox{\columnwidth}{!}{%
\begin{tabular}{ll ccc}
\toprule
\textbf{Pair} & \textbf{Task} & \textbf{DLM RDI} & \textbf{AR RDI} & \textbf{Adv.} \\
\midrule
\multirow{3}{*}{LLaDA} & TriviaQA & 0.275 & 0.512 & $+0.237$ \\
                       & GSM8K    & 0.488 & 0.527 & $+0.039$ \\
                       & ARC      & 0.100 & 0.181 & $+0.081$ \\
\midrule
\multirow{3}{*}{Dream} & TriviaQA & 0.222 & 0.333 & $+0.111$ \\
                       & GSM8K    & 0.383 & 0.460 & $+0.077$ \\
                       & ARC      & 0.052 & 0.053 & $+0.001$ \\
\bottomrule
\end{tabular}%
}
\end{table}

\section{Calibration Confidence Aggregation}
\label{app:calib_ablation}

To calculate DLM sequence-level confidence, we use the geometric mean of per-token probabilities (which equates to the exponential of the mean per-token log-probability). Under a log-calibration view, proper scoring rules for sequences decompose additively in log-space; therefore, the geometric mean is the only aggregation method that preserves comparability across different sequence lengths. We apply this exact aggregation to both \dlm{} and \ar models across all calibration metrics to ensure strict like-for-like comparability.

\paragraph{Confidence-definition ablation.}
Table~\ref{tab:calib_ablation} reports mean ECE (over TriviaQA/ARC/GSM8K, $n{=}200$) under alternative per-token bases (maximum vs.\ chosen-token vs.\ commitment-time probability) and aggregations (geometric mean vs.\ arithmetic mean vs.\ minimum). The overconfidence ordering (\dlm{} $\gg$ \ar) survives every non-degenerate definition: \dlm ECE stays $0.30$--$0.43$ versus \ar $0.19$--$0.21$. For greedy \ar decoding, max $=$ chosen $=$ commit (a consistency check). The only exception is \texttt{min} aggregation, which is degenerate---it collapses a sequence to its single least-confident token and drives the \ar models \emph{under}-confident (ECE up to $0.567$)---so it does not reflect \dlm calibration. Selective-prediction AUROC stays $0.62$--$0.77$ for the \dlm{}s across all definitions, so the ranking argument is definition-stable.

\begin{table}[htbp]
\centering
\small
\caption{Confidence-definition ablation: mean ECE over TriviaQA/ARC/GSM8K ($n{=}200$). Left columns vary the per-token basis (aggregation fixed to geometric mean); right columns vary the aggregation (basis fixed to max probability). ``---'': Dream's sampler exposes no distinct commitment-time probability.}
\label{tab:calib_ablation}
\resizebox{\columnwidth}{!}{%
\begin{tabular}{l ccc cc}
\toprule
& \multicolumn{3}{c}{\textbf{Per-token basis (geo)}} & \multicolumn{2}{c}{\textbf{Aggreg.\ (max)}} \\
\cmidrule(lr){2-4}\cmidrule(lr){5-6}
\textbf{Model} & max & chosen & commit & mean & min \\
\midrule
LLaDA-8B (DLM)   & 0.433 & 0.390 & 0.385 & 0.434 & 0.360 \\
LLaMA-3-8B (AR)  & 0.187 & 0.187 & 0.187 & 0.211 & 0.567 \\
Dream-7B (DLM)   & 0.319 & 0.298 & ---   & 0.321 & 0.335 \\
Qwen2.5-7B (AR)  & 0.189 & 0.189 & 0.189 & 0.203 & 0.322 \\
\bottomrule
\end{tabular}%
}
\end{table}

\section{Extended Limitations Discussion}
\label{app:limitations}

\paragraph{Tokenizer Mismatch.}
LLaDA uses the LLaMA-3 tokenizer, which was not explicitly designed for masked prediction. It may tokenize certain perturbations (e.g., homoglyphs) differently than a purely \dlm{}-native tokenizer would. This could theoretically inflate the measured robustness degradation for homoglyph perturbations.

\paragraph{Generation Length Sensitivity.}
\dlm{}s fix their generation length at $T$ steps with $L$ masked positions. We cap \texttt{max\_new\_tokens} at 128 (or 512 for GSM8K) across our experiments. Tasks requiring substantially longer outputs (e.g., complex code generation) may exhibit different robustness profiles, as the iterative commitment process has more steps over which to accumulate noise-induced errors.

\paragraph{Inter-Rater Reliability of Answer Matching.}
For TriviaQA, we evaluate using substring matching against all provided aliases. While standard, this method can overcount correct answers for short aliases and undercount paraphrased correct answers. Future work extending these findings could benefit from model-based evaluation paradigms (e.g., BERTScore~\citep{bertscore2019}).

\section{Perturbation Examples}
\label{app:perturbation_examples}

Table~\ref{tab:pert_examples} illustrates representative examples of each perturbation type applied at a rate of $r{=}0.15$.

\begin{table}[htbp]
\centering
\small
\caption{Representative perturbation examples at noise rate $r{=}0.15$.}
\label{tab:pert_examples}
\resizebox{\columnwidth}{!}{
\begin{tabular}{p{2.5cm} p{4.5cm}}
\toprule
\textbf{Type} & \textbf{Example} \\
\midrule
Original & \textit{What is the capital of France?} \\
\midrule
Transposition & \textit{Wha tis hte cpaital of Farce?} \\
Deletion & \textit{What is the apital of rance?} \\
Keyboard & \textit{Whay is the capiral of France?} \\
Homoglyph & \textit{Whаt is thе cарitаl оf Frаnce?} \\
\midrule
Word deletion & \textit{What is the of France?} \\
Local shuffle & \textit{What is capital the of France?} \\
\bottomrule
\end{tabular}
}
\end{table}

\section{AI Assistance Declaration}
\label{app:ai_usage}

During the preparation of this manuscript, the authors utilized an AI assistant to refine sentence structure, improve readability, and polish the academic prose. The AI was not used to generate novel scientific claims, design experiments, or analyze data. The authors carefully reviewed and edited all AI-refined text and take full responsibility for the final content of this paper.

\end{document}